\documentclass[nohyperref]{article}

\usepackage{microtype}
\usepackage{graphicx}
\usepackage{subfigure}
\usepackage{booktabs} 

\usepackage{hyperref}

\usepackage[accepted]{icml2022}

\usepackage{amsmath}
\usepackage{amssymb}
\usepackage{mathtools}
\usepackage{amsthm}

\usepackage[capitalize,noabbrev]{cleveref}

\theoremstyle{plain}

\theoremstyle{definition}

\theoremstyle{remark}

\usepackage[textsize=tiny]{todonotes}
\usepackage{enumerate}
\usepackage{multirow}
\usepackage{algorithm}
\usepackage{algorithmic}
\usepackage{tabularx}
\usepackage{xcolor,colortbl}
\usepackage{nicefrac}
\usepackage{booktabs}
\usepackage{makecell}
\usepackage{xspace}
\usepackage{graphbox}
\usepackage{pifont}

\icmltitlerunning{
On the interplay of adversarial robustness and architecture components: patches, convolution and attention
}

\newcommand{\cls}{\texttt{CLS}\xspace}

\newcommand{\ffcv}{\texttt{FFCV}\xspace}
\newcommand{\torchvision}{\texttt{torchvision}\xspace}

\definecolor{Gray}{gray}{0.85}
\definecolor{LightCyan}{rgb}{0.88,1,1}

\newcolumntype{C}[1]{>{\centering\arraybackslash}p{#1}}
\newcolumntype{L}[1]{>{\raggedright\arraybackslash}p{#1}}
\newcolumntype{R}[1]{>{\raggedleft\arraybackslash}p{#1}}

\newcommand{\norm}[1]{\left\|#1\right\|}

\begin{document}

\twocolumn[
\icmltitle{
           
           On the interplay of adversarial robustness and architecture components: patches, convolution and attention
           }

\icmlsetsymbol{equal}{*}

\begin{icmlauthorlist}
\icmlauthor{Francesco Croce}{tue}
\icmlauthor{Matthias Hein}{tue}
\end{icmlauthorlist}

\icmlaffiliation{tue}{University of T{\"u}bingen, Germany}

\icmlcorrespondingauthor{Francesco Croce}{francesco.croce@uni-tuebingen.de}

\icmlkeywords{Machine Learning, ICML}

\vskip 0.3in
]

\printAffiliationsAndNotice{}  

\begin{abstract}
In recent years novel architecture components for image classification have been developed, starting with attention and patches used in transformers. While prior works have analyzed the influence of some aspects of architecture components on the robustness to adversarial attacks, in particular for vision transformers, the understanding of the main factors is still limited. We compare several (non)-robust classifiers with different architectures and study their properties, including the effect of adversarial training on the interpretability of the learnt features and robustness to unseen threat models. An ablation from ResNet to ConvNeXt 
reveals key architectural changes leading to almost $10\%$ higher $\ell_\infty$-robustness.
\end{abstract}

\section{Introduction}
The introduction of vision transformers (ViTs) \citep{dosovitskiy2021an} showed that different architectures can perform on par or even better than convolutional networks (CNNs) in various computer vision tasks. This led to active research on optimizing the network design for better performance e.g. classification accuracy on the ImageNet dataset \citep{deng2009imagenet}, which in turn resulted in several new architectures 
\citep{touvron2021training,liu2021Swin, trockman2022patches, liu2022convnet}. It is however still partially unexplained which are the key components which make an architecture effective in a specific task \citep{park2022how}.
It has been noticed that some architectures might natively perform better than others in side tasks, e.g. \citet{fort2021exploring} argue that ViTs have significantly better out-of-distribution detection ability. Moreover, recent works \citep{bhojanapalli2021understanding, paul2021vision} suggest that ViTs are more robust to common corruptions in ImageNet-C \citep{HenDie2019}. 
About robustness to adversarial attacks, \citet{shao2021adversarial} suggest that naturally trained ViTs are more robust to $\ell_\infty$-bounded perturbations 
than ResNets. 
Conversely, ViTs appear more vulnerable to patch attacks \citep{gu2021vision}, while the comparison is mixed on $\ell_0$-attacks \citep{fu2022patchfool}. However, \citet{bai2021are} reports that such differences might disappear with either adversarial training \citep{MadEtAl2018} for $\ell_\infty$ or using similar training protocols for the two architectures. Finally, \citet{debenedetti2022adversarially} has recently achieved SOTA results for the $\ell_\infty$-threat model on ImageNet using XCiT \citep{el2021xcit}, a transformer-like network which reintroduces convolutions as part of its architecture.

In this work, we first extend the analysis of the robustness of different normally trained architectures to adversarial patches: unlike previous attacks, we show that on networks which divided the input image in disjoint tokens e.g. ViTs it is preferable to position the adversarial patch to cover multiple tokens instead of a single one, although this might yield lower loss values in the case of traditional vision transformers. However, this phenomenon is largely mitigated when considering robust models. Then, we explore how the features learnt by classifiers when using adversarial training wrt $\ell_\infty$ differ from those of plain models, showing e.g. how the attention maps of ViTs gain in interpretability. Moreover, 
we study how robustness wrt $\ell_\infty$, 
generalizes to unseen attacks, both $\ell_p$-bounded 
and not. While ResNets generally attains worse generalization, we show in an extensive ablation study regarding the transition from a ResNet to the ConvNeXt \cite{liu2022convnet} architecture that small modifications in the architecture are sufficient to, at least partially, close the gap to ViTs. 
robust ConvNeXt 
which achieves $46.2\%$ robust accuracy against $l_\infty$-perturbations of $\epsilon=4/255$ outperforming the recent $41.7\%$ for the XCiT architecture \citet{debenedetti2022adversarially} and also being about $10\%$ higher than a ResNet-50 architecture. However, even more interesting is that a relatively small change of the traditional ResNet-50 architecture achieves $44.0\%$ robust accuracy. 

\section{Background and related works} \label{sec:background}
In this section we provide the necessary background on the architectures we consider and adversarial robustness.
\subsection{Architectures} \label{sec:models}

\textbf{ResNets:} ResNet \citep{he2016identity} and WideResNet \citep{ZagKom2016} rely on convolutions and on the usage of residual connections, with non-linearity provided by activation functions as ReLU and GELU \citep{hendrycks2016gaussian}. Such models held for long time SOTA results on vision tasks, and are still dominant when considering adversarial robustness 
\citep{croce2020robustbench} for CIFAR-10 and ImageNet. ResNets 
do not divide the input image in disjoint patches.

\textbf{Vision transformers:} \citet{dosovitskiy2021an} introduced a convolution-free architecture for vision applications inspired by the transformer models used in natural language processing. ViTs split the input image in disjoint patches, in analogy of the language tokens, add a class token (\cls) used for classification, then process them with blocks including multi-head self-attention \citep{vaswani2017attention} and MLPs. Later, \citet{touvron2021training} showed how it is possible to train ViTs efficiently: we use their models named DeiT with $16 \times 16$ patches. 

\textbf{Cross-covariance vision transformers:} \citet{el2021xcit} replaced in the ViT design the traditional (local) self-attention with the so-called cross-covariance version of it (XCA). Moreover, each block contains two depth-wise convolutional layers to allow better communication among patches. 
The resulting XCiT compares favorably to DeiT with respect to performance in classification and segmentation tasks and memory usage.

\textbf{ConvNeXt:} To close the gap between CNNs and ViTs on ImageNet, \citet{liu2022convnet} modify the ResNet backbone to make it more similar to the SOTA Swin transformer \citep{liu2021Swin}, until they outperform it. 
ConvNeXt adopts patchified stem, i.e. non overlapping convolutions in the first layer, and depth-wise convolutions. 

For our analysis, we use
ResNet-50, 
WideResNet-50-2, 
DeiT-S 
XCiT-S, 
ConvNeXt-T, 
which have, except for the WideResNet, comparable size in terms of number of parameters. 
All are trained on the ImageNet-1k dataset 
and use image resolution $224\times 224$ pixels.
For naturally trained classifiers we use the checkpoints provided by either \torchvision model zoo or the \texttt{timm} library \citep{rw2019timm}, while the robust ones are made available by prior works (ResNet-50 and DeiT-S are from \citet{bai2021are}, WideResNet-50-2 from \citet{salman2020adversarially}, XCiT-S from \citet{debenedetti2022adversarially}). 

\subsection{Adversarial robustness}
The output of a neural network can be easily 
modified by small perturbations of an input which do not change its semantic content \citep{BigEtAl13, SzeEtAl2014}. 
Many works have focused on developing methods
to find such adversarial perturbations whose size is constrained by a certain metric, e.g. an $\ell_p$-norm \citep{CarWag2016}, or which are limited to have a specific shape like square patches \citep{karmon2018lavan} or frames \citep{zajac2019adversarial}. 
At the same time, 
the most successful method to obtain adversarially robust models is 
adversarial training \citep{MadEtAl2018}, 
and its recent advances \citep{ZhaEtAl2019,CarEtAl19, gowal2020uncovering, rebuffi2021data}.
\citet{gu2021vision, fu2022patchfool, lovisotto2022give} have suggested that ViTs, 
are more vulnerable to adversarial patches than ResNets. 
Similarly \citet{shao2021adversarial} argued that ViTs are more robust than other architectures in the $\ell_\infty$-threat model, 
However, \citet{bai2021are} report that, when trained with similar augmentations, DeiTs and ResNets have similar robustness to adversarial patches, and using GELU in the ResNet backbone suffices to obtain with adversarial training classifiers as robust as DeiTs with respect to $\ell_\infty$-perturbations with size $\epsilon=4/255$ on ImageNet.

\section{Robustness of natural and robust models to adversarial patches} \label{sec:patches}

In this section we analyze the interaction between patch attacks and the token grid used by DeiT and XCiT, 
Then, by developing a simple greedy attack, we show that both ViTs and ResNet are less robust to adversarial patches than what has been reported by previous works.

\subsection{Effect of grid aligned and non-grid aligned patches on adversarial loss}
Both \citet{fu2022patchfool,gu2021vision} evaluate the robustness of vision transformers using mostly patches which are aligned with the grid of the input tokens, so that one adversarial patch exactly covers one cell of the tokens grid. However, adversarial attacks might benefit from modifying multiple tokens by a smaller fraction: we consider patches which are centered at the intersection point of the tokens grid (each patch covers 1/4 of 4 contiguous tokens). For transformers with $16\times 16$ tokens (and using patches of the same size) this yields $14\times 14 = 196$ possible positions of \textit{aligned} patches and $13 \times 13 = 169$ \textit{non aligned} ones. For each one we maximize the margin loss \citep{CarWag2016} for 100 iterations with APGD \citep{croce2020reliable}. In Fig.~\ref{fig:loss_maps} we show the results 
for models trained either naturally (left) or with adversarial training wrt $\ell_\infty$ (right). Among the naturally trained models, for DeiT-S the grid-aligned patches attain much higher loss values than the non aligned ones, while both yielding 0\% robust accuracy (see below). Conversely, there is no particular difference when using ResNet-50 or XCiT-S, which also use convolutional layers. 
The adversarially trained DeiT-S does not show the same behavior 
of the plain one. Moreover, for all adversarially trained architectures, the attacks achieve higher loss at patch locations in correspondence of class-specific features in the image, creating a sort of saliency map, in strong contrast to non robust classifiers. This is in line with the observation that robust models have gradients significantly aligned with human perception \citep{TsiEtAl18,santurkar2019image}.

\newlength{\newl}
\setlength{\newl}{.55\columnwidth}
\begin{figure*}[t] \centering \small
\tabcolsep=2pt
\begin{tabular}{c| ccc | ccc}
\multirow{3}{*}{\textit{image}}& \multicolumn{3}{c|}{\textit{plain training}} & \multicolumn{3}{c}{\textit{adversarial training}} \\[1mm]
& DeiT-S &   XCiT-S & ResNet-50 & DeiT-S  & XCiT-S& ResNet-50\\[1mm]
& \hspace{.5mm} A \hspace{5mm} NA & \hspace{.5mm} A \hspace{5mm} NA & \hspace{.5mm} A \hspace{5mm} NA & \hspace{.5mm} A \hspace{5mm} NA & \hspace{.5mm} A \hspace{5mm} NA & \hspace{.5mm} A \hspace{5mm} NA\\
\toprule
\includegraphics[height=\newl]{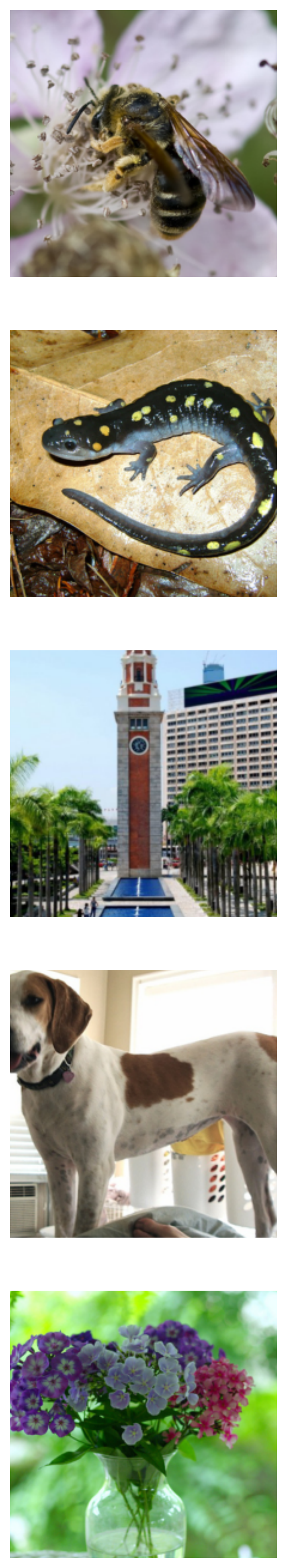}&
\includegraphics[height=\newl]{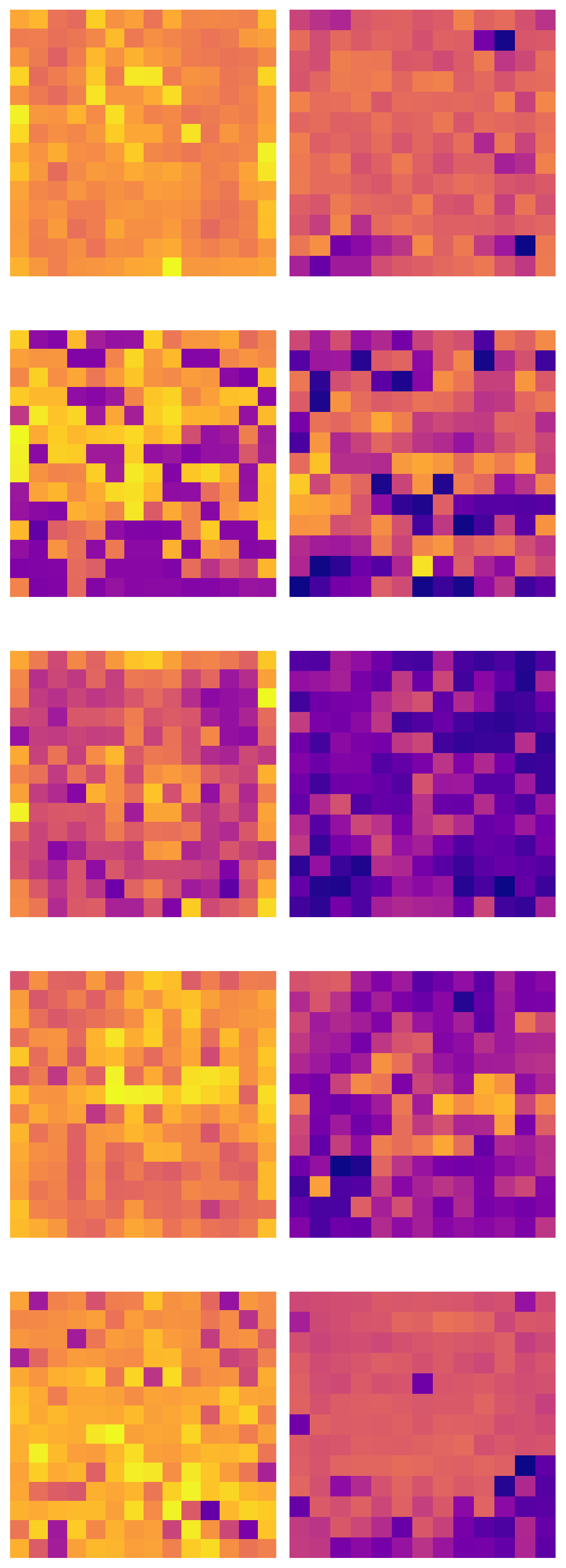}&
\includegraphics[height=\newl]{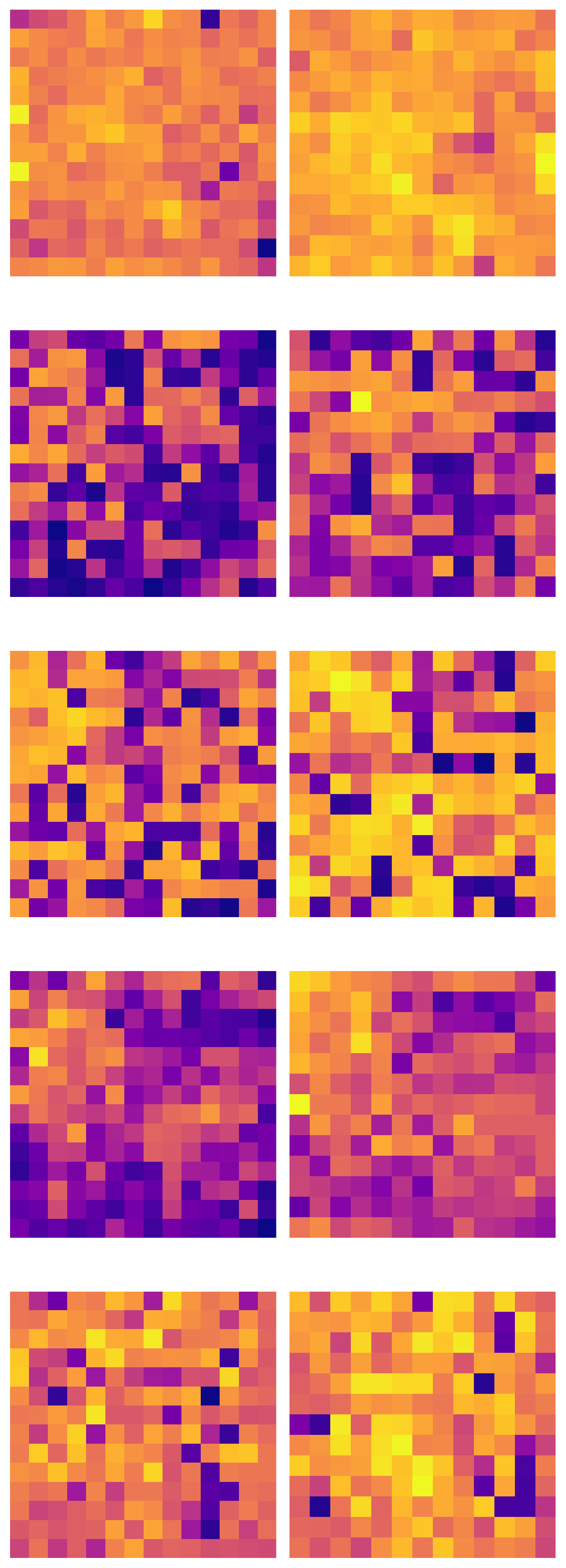} &
\includegraphics[height=\newl]{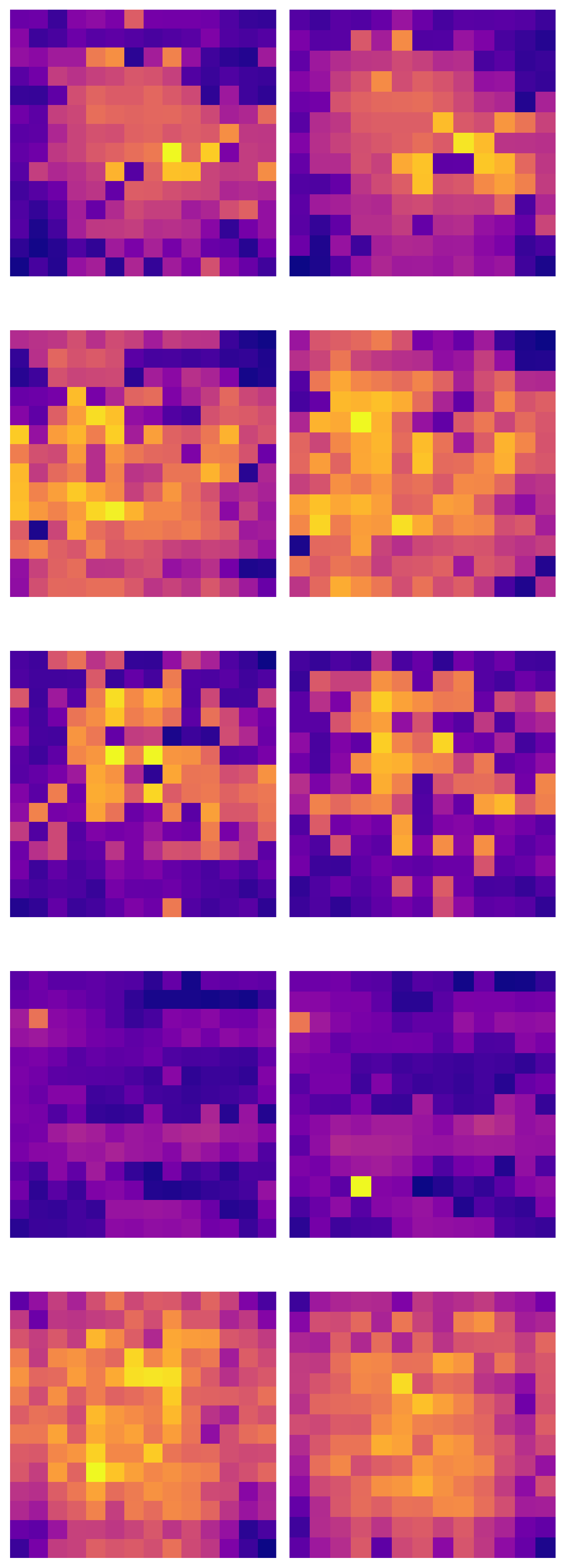}&
\includegraphics[height=\newl]{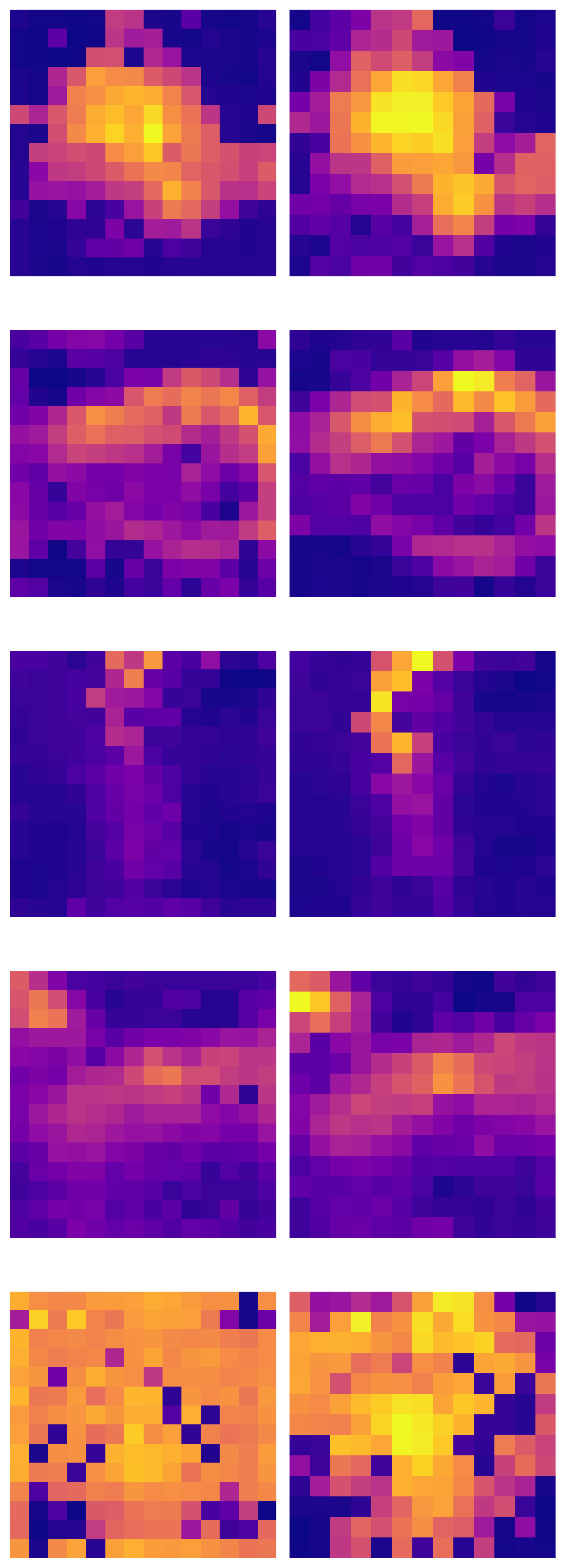} &
\includegraphics[height=\newl]{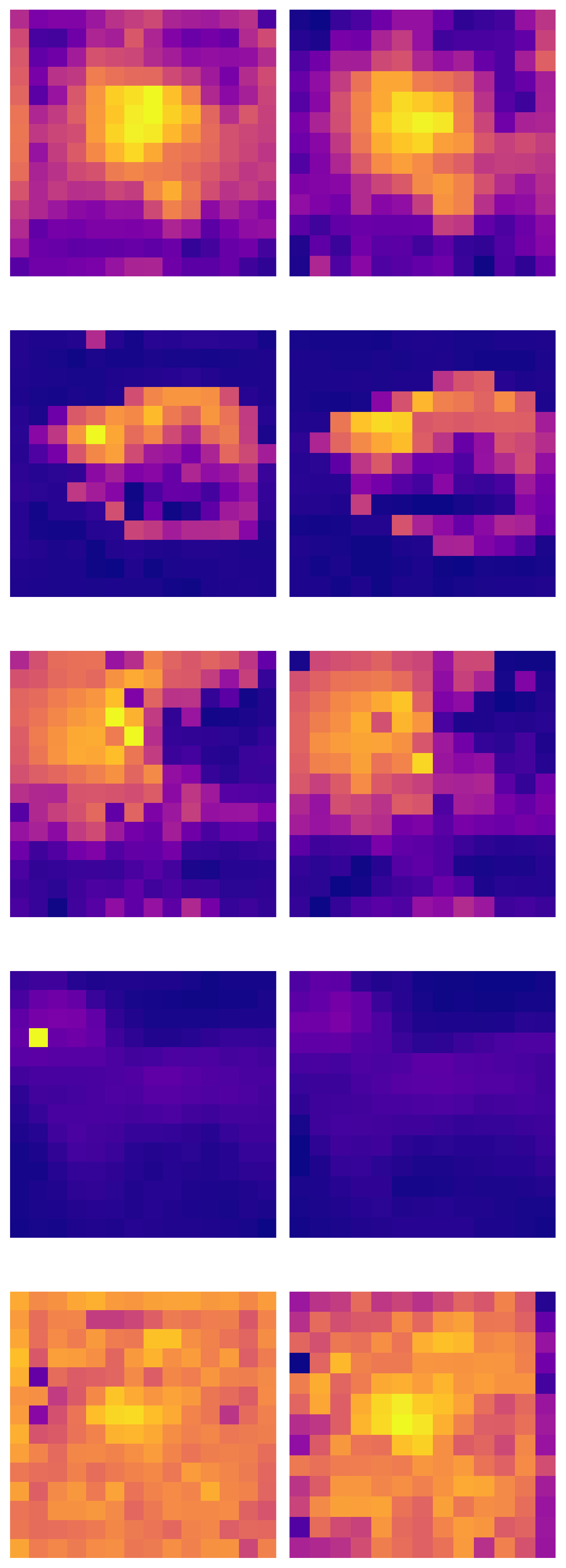}&
\includegraphics[height=\newl]{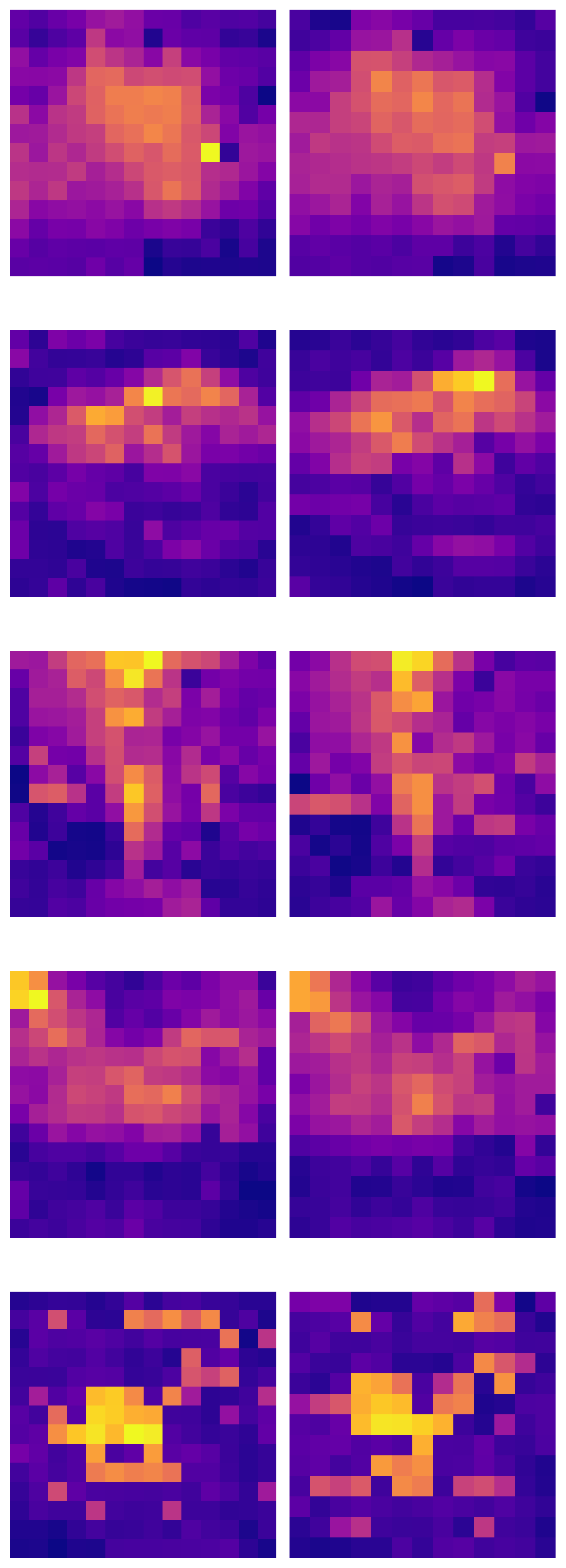}
\end{tabular}\caption{
For each image and model we show the map of the values obtained maximizing the margin loss within disjoint patches of size $16\times 16$ pixels which are either aligned (A) with the grid of the tokens of ViTs or not (NA) i.e. centered on the intersection the grid (that is having a $4 \times 4$ overlap with the four adjacent patches). Each pair (aligned and not) of plots is normalized independently. Brighter colors indicate higher values of the loss.
} \label{fig:loss_maps}
\end{figure*}

\subsection{Robustness of plain and robust classifiers to adversarial patches}
\textbf{A simple patch attack:} several prior works \citep{BroEtAl2017, yang2020patchattack, croce2020sparsers} have introduced patch attacks. 
Here we adopt a more greedy strategy which allows us to compare the effectiveness of patches aligned or not with the token grid. For both aligned patches and not aligned one, we optimize the margin loss for 20 iterations with APGD 
($\ell_\infty$-version, with initial step size 0.5) for all possible patches 
then we select the 20\% of patches of each type with highest loss and optimize their content for 480 iterations 
(the location of the patches remains unchanged). 

\textbf{Robustness to aligned and non aligned patches:} Table~\ref{tab:robustness_aligned_and_not_patches} shows the robust accuracy to aligned and not aligned patches, together with the worst-case over them. 
First we analyze the plain models: for ResNet-50 the aligned and non-aligned perform similarly as expected since there is no tokenization. On DeiT-S both perform equally well, despite the aligned patches achieving significantly higher loss values on average. 
Finally, for XCiT-S, 
the non aligned patches achieve higher success rate than the aligned ones. 
Overall, XCiT-S appears more robust than both DeiT-S and ResNet-50, implying that transformer-based architectures are not necessarily more vulnerable to adversarial patches. 
Moreover, our greedy attack yields much lower robust accuracy for both DeiT-S and ResNet-50 compared to what has been reported by \citet{fu2022patchfool}, that is 6.25\% and 24.00\% respectively (although different subsets of test points are used). 
For adversarially trained classifiers wrt the $\ell_\infty$-threat model, the robust accuracy for all architectures against patch attacks improves, with DeiT-S being the most robust one, and there is no clear trend regarding the comparison of aligned and non-aligned patches. 

\begin{table}[h]
    \centering \small
    \tabcolsep=1.5pt
    \caption{Robust accuracy (\%) of different ImageNet models (computed for 1000 images)  to patch perturbations which are either aligned or non-aligned with the tokens grid (for ResNet-50 the grid of the transformers is used). Worst-case is over both 
    patches.} 
    \label{tab:robustness_aligned_and_not_patches}
    \begin{tabular}{C{14mm}|*{3}{C{10mm}} |*{3}{C{10mm}}}
         \multirow{2}{*}{\textit{patch type}}&  
         \multicolumn{3}{c|}{\textit{plain training}} & \multicolumn{3}{c}{\textit{adversarial training}} \\[1mm]
& DeiT-S &   XCiT-S & RN-50 & DeiT-S  & XCiT-S& RN-50
         \\
         \toprule

        aligned & 0.0 & 6.7 & 1.7 & 24.7 & 20.4 & 15.3 \\
not align. & 0.0 & 4.9 & 1.8 & 22.5 & 20.1 & 17.4 \\ \midrule
worst-case & 0.0 & 4.1 & 1.2 & 21.4 & 18.7 & 14.5
        \\ \bottomrule
    \end{tabular}
\end{table}

\textbf{Robust to patches at a fixed-position:} \citet{lovisotto2022give} have recently reported that DeiT-B, the larger version of DeiT-S, has a non trivial robustness of 13.1\% when using a $16\times 16$ patch at a fixed position, i.e. in the top left corner. However, we managed, by using our attack which optimizes the margin loss with APGD, to reduce it to 0\%.

\section{Effect of adversarial training on ViTs features}

We now study how the interpretability of the representations learnt by vision transformers is affected by adversarial training, with  $\ell_\infty$-threat model with radius $\epsilon=4/255$. 

\subsection{Interpretability of attention maps}
In Fig.~\ref{fig:attn_maps} we show for the DeiT-S classifiers the attention maps of the \cls token for each head in the last block: in each row, the first 6 maps are produced with the plain model, while the remaining 6 are from the robust one, all corresponding to the original image shown on the far left. 
The analogous maps for XCiT 
are in Fig.~\ref{fig:attn_maps_xcit} in appendix. In both cases, the maps of the robust models are significantly more interpretable than those of the plain ones: the various heads are triggered by different parts (and objects) of the image, unlike those from the plain classifier which are mostly sparse and similar to each other. Interestingly, this is similar to what happens to the attention maps of models trained by self-supervision with DINO \citep{caron2021emerging}.

\setlength{\newl}{.55\columnwidth}
\begin{figure*}[t] \centering \small
\tabcolsep=4.5pt
\begin{tabular}{c | c | c}

\textit{original} & \textit{plain training} & \textit{adversarial training}\\ \toprule
\includegraphics[height=\newl]{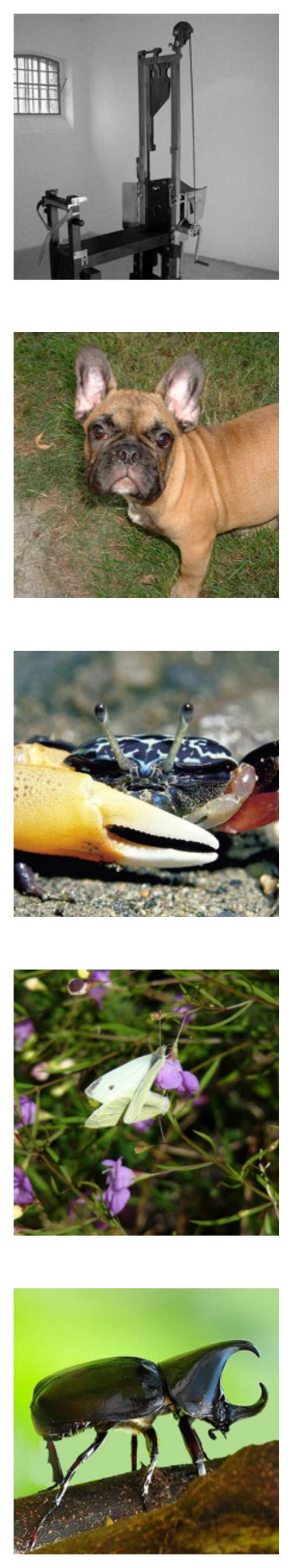}&
\includegraphics[height=\newl]{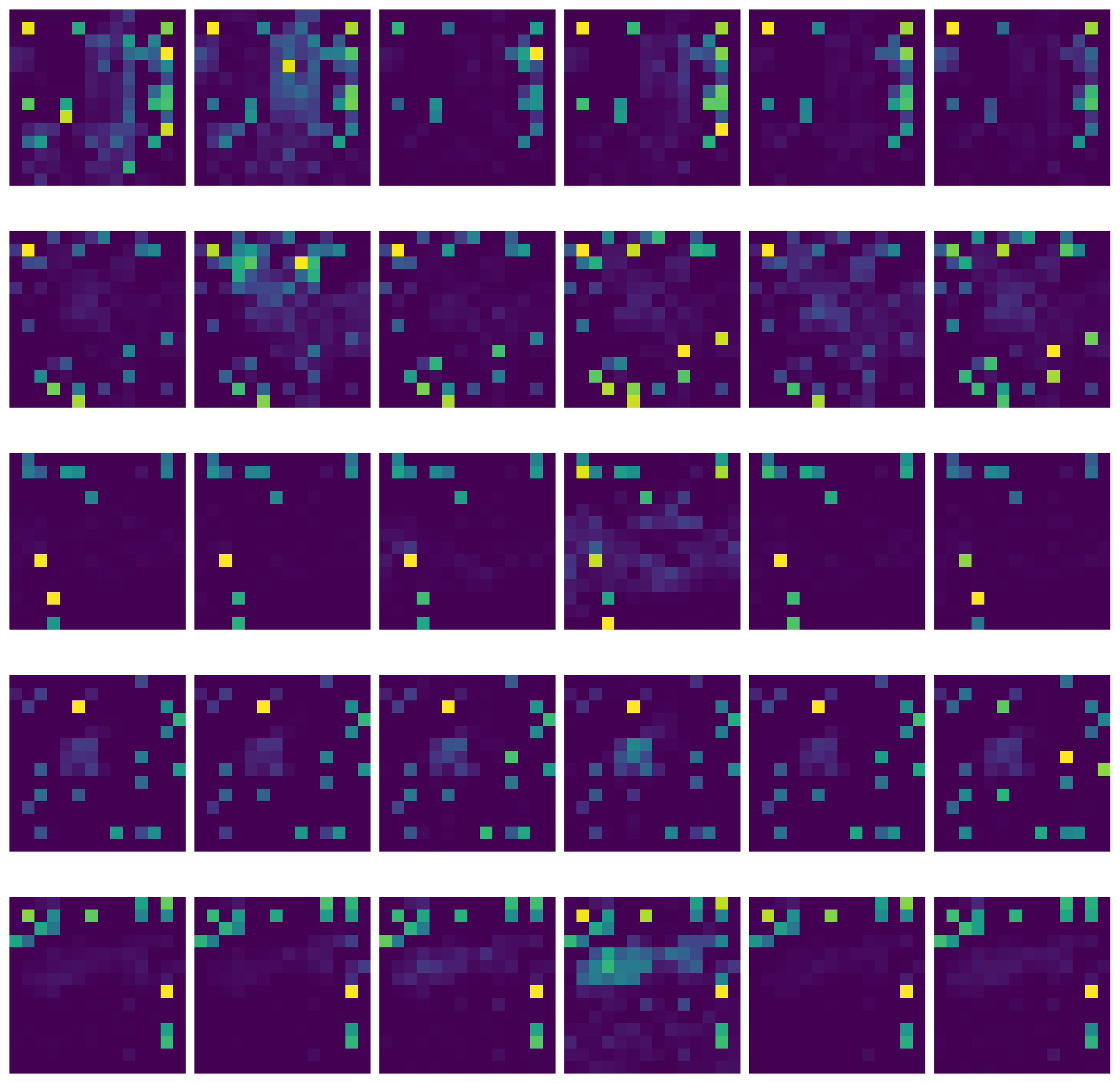}&
\includegraphics[height=\newl]{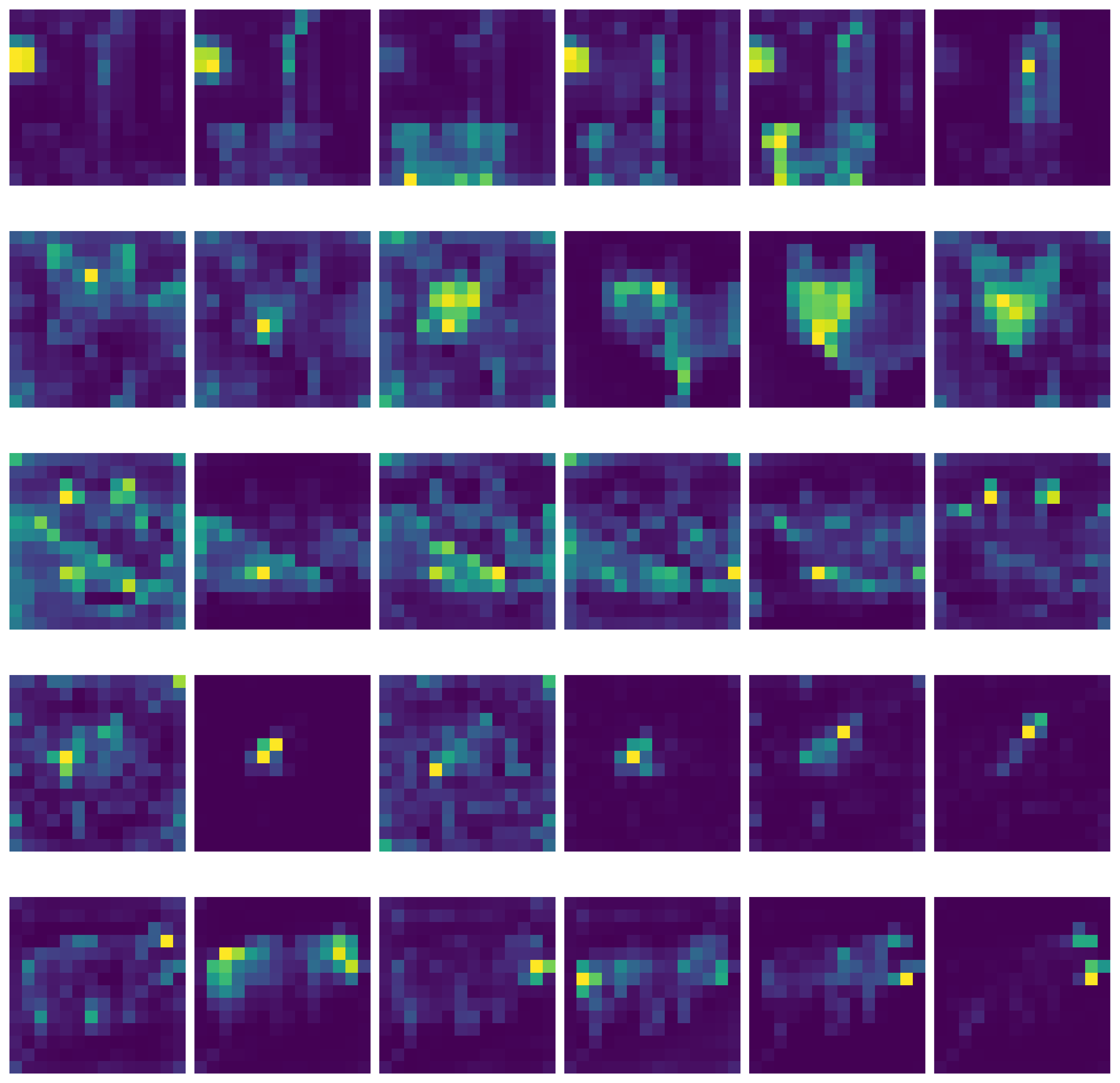}\\

\end{tabular}\caption{Attention maps of the \cls token for each head of the last layer of a DeiT-S model 
The attention of the robust DeiT-S is concentrated on the object, where each head pays attention to different parts.}
\label{fig:attn_maps}\end{figure*}

\subsection{Inner representations of XCiT} \label{sec:inner_repr_xcit}

\citet{el2021xcit} noticed that the norm across the features dimension of queries and keys of each token may indicate the salient regions in the image (higher values correspond to more salient areas). We show this for the plain model in the top part of Fig.~\ref{fig:keys_heads_xcit} for the keys of the last XCA block (each column corresponds to a head), using images of resolution $384 \times 384$. While these identify meaningful features, they also show highly activated patches at random positions. This effect is instead absent for the robust model (bottom part of Fig.~\ref{fig:keys_heads_xcit}), whose images appear ``denoised'', with different heads focusing on complementary details of the image. We plot the maps of the robust model for three images containing dogs in Fig.~\ref{fig:xcit_keys_class_specific}: 
the heads focus on similar features across images 
(see also the appendix).

\setlength{\newl}{.45\columnwidth}
\begin{figure*}[h] \centering \small
\begin{tabular}{c c}
\multicolumn{2}{c}{\textbf{standardly trained XCiT-S}}
\\
\includegraphics[height=\newl]{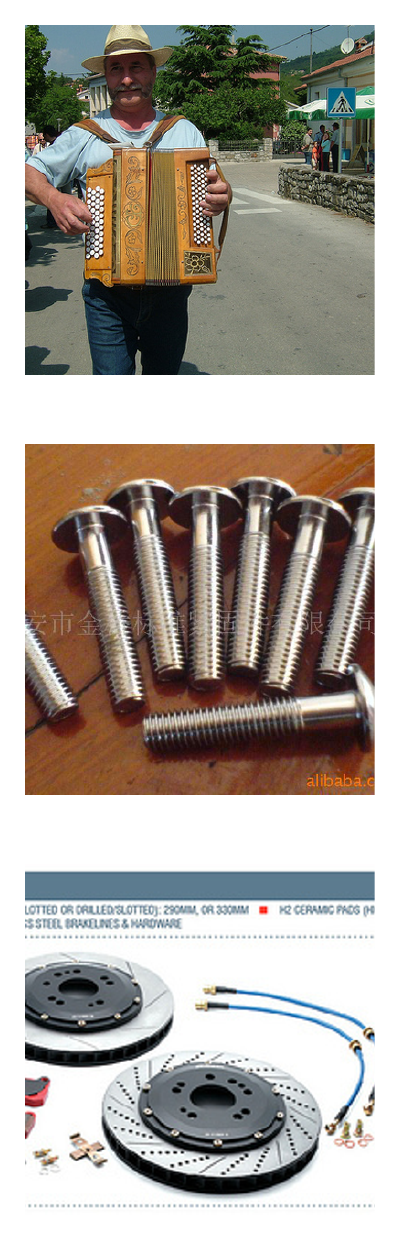} &
\includegraphics[height=\newl]{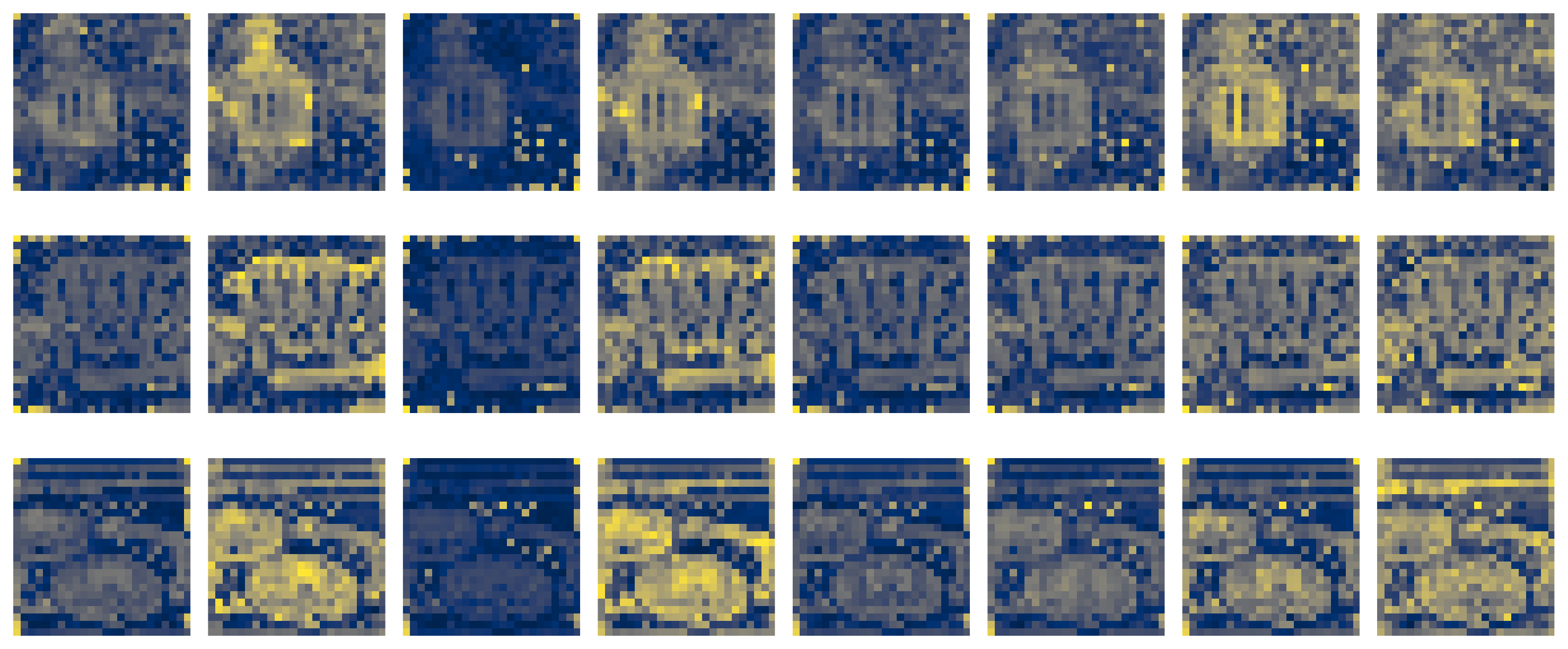}
\\
\multicolumn{2}{c}{\textbf{adversarially trained XCiT-S}}\\
\includegraphics[height=\newl]{figures/pl_orig_xcit_ks.pdf} &
\includegraphics[height=\newl]{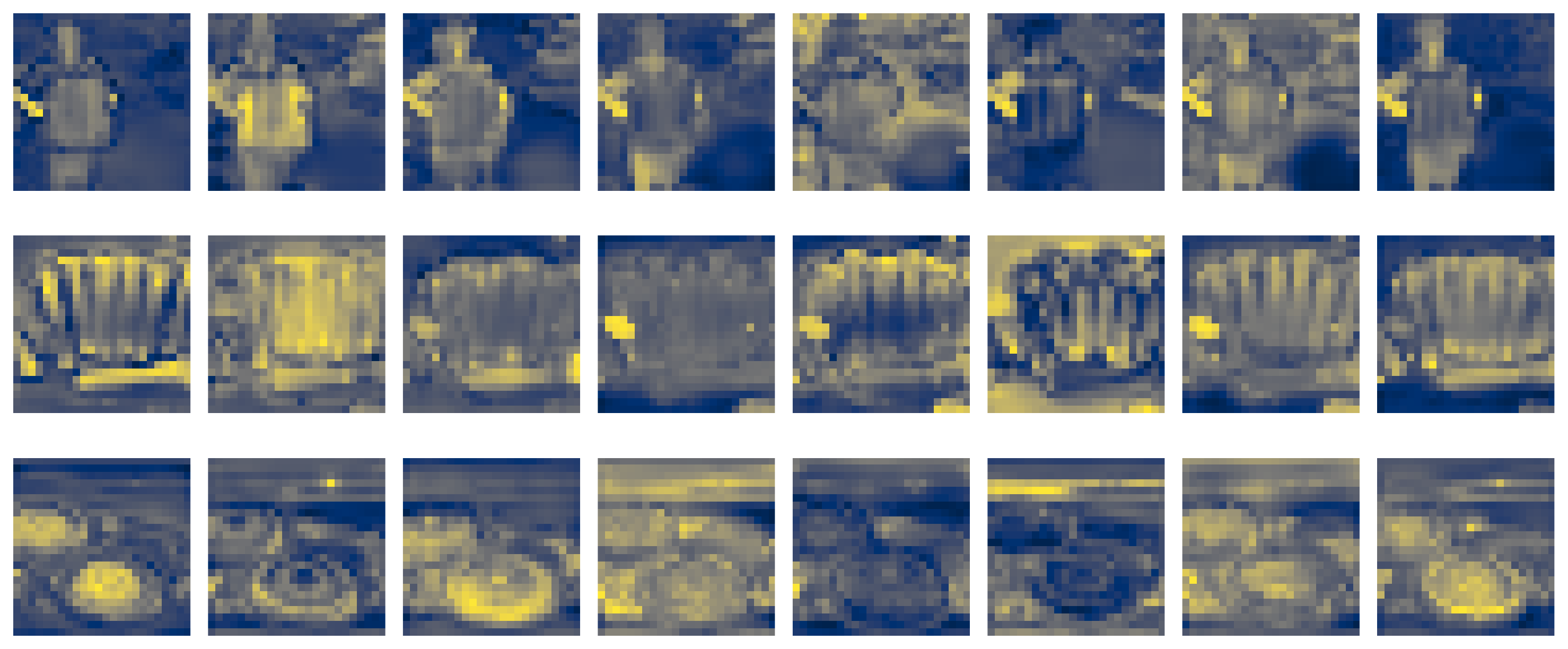}
\end{tabular}\caption{Visualization of the norm across the features dimension of the keys of the last XCA block, with each column representing a head, for a normally trained model (top) and an adversarially trained model (bottom). Images are with resolution $384\times 384$.} \label{fig:keys_heads_xcit}
\end{figure*}

\setlength{\newl}{.45\columnwidth}
\begin{figure*}[h] \centering \small
\begin{tabular}{c c}
\multicolumn{2}{c}{\textbf{}} \\
\includegraphics[height=\newl]{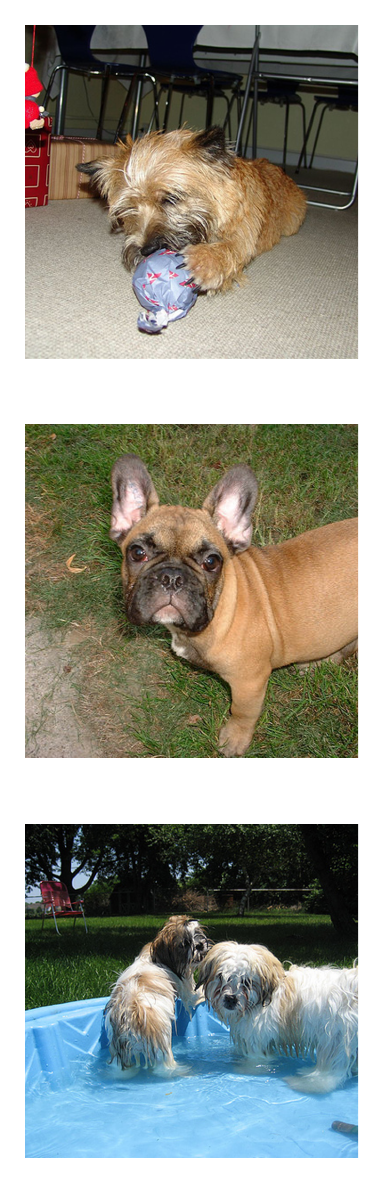} &
\includegraphics[height=\newl]{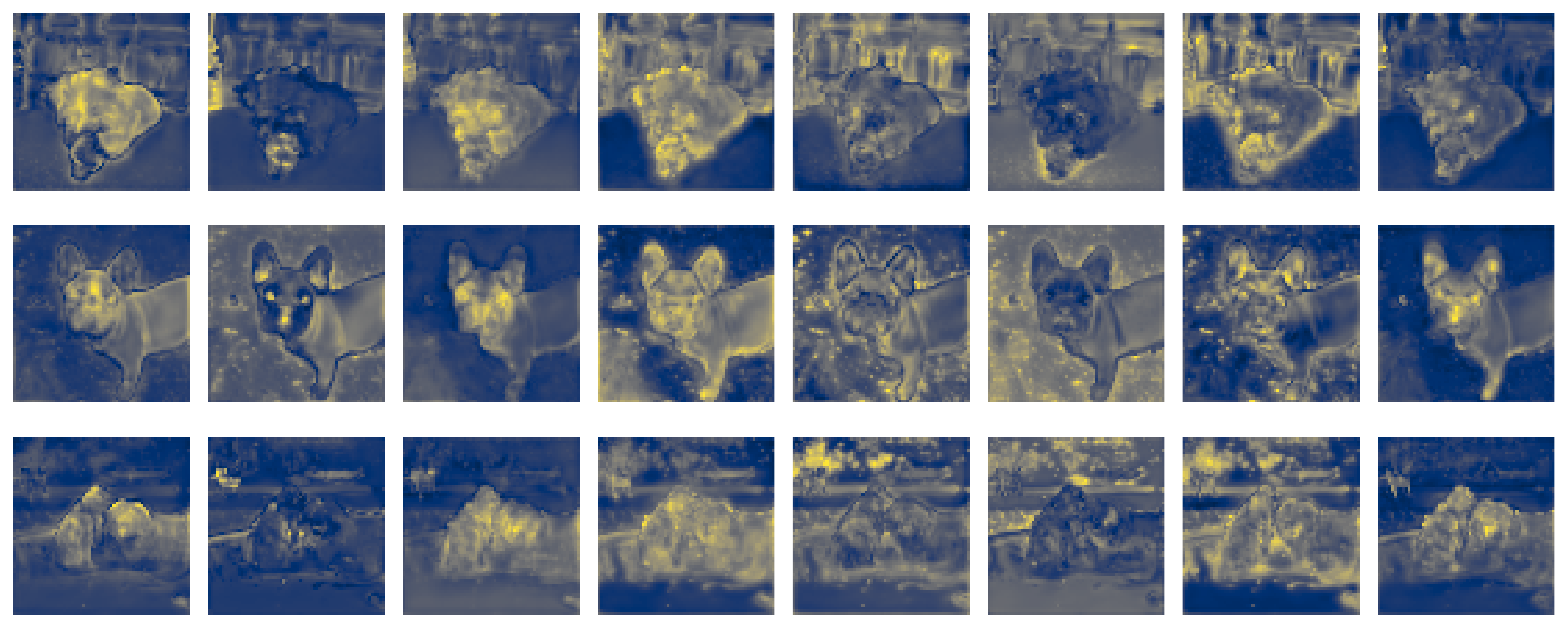}
\end{tabular}
\caption{Comparison of the visualization of norm across the features dimension of the keys of the last XCA block of the robust XCiT-S for different images of resolution $1024\times 1024$ of similar classes: similar features of the dog are highlighted across images.} \label{fig:xcit_keys_class_specific}
\end{figure*}

\section{Generalization of robustness to unseen threat models} \label{sec:generalization_robustness}
It is known that robustness in a specific threat model does not necessarily generalizes to other ones, e.g. across $\ell_p$-norms \citep{TraBon2019,kang2019transfer}. We study here  to which extent this varies across architectures, and test how the classifiers with adversarial training wrt $\ell_\infty$ behave in other, unseen, threat models. This might hint to which architectural components are relevant for designing more robust models. 
In fact, the results reported in Sec.~\ref{sec:patches} already suggest that not all network trained to be robust wrt $\ell_\infty$ are equally vulnerable to adversarial patches. 
Moreover, this allows to check whether some architectures are preferable wrt the robustness in the threat model used for training i.e. $\ell_\infty$.

\textbf{Experimental setup:} We report all statistics on 1000 points. 
For the $\ell_\infty$, $\ell_2$ and $\ell_1$ threat models we use bounds 4/255, 2 and 75 respectively, and APGD for cross-entropy and DLR-loss from AutoAttack \citep{croce2020reliable} as attacks. We test robustness wrt $\ell_0$ in pixel space (that is all color channels of a selected pixels are perturbed) for bound $\epsilon_0=100$ 
using \texttt{Sparse-RS} \citep{croce2020sparsers} with 50,000 queries, after running the white-box but less effective PGD\textsubscript{0} \citep{croce2019sparse} to quickly reduce the points to test. For adversarial patches we keep the same setup as used in Sec.~\ref{sec:patches}, i.e. $16\times 16$ perturbations, with our greedy attack. Finally, we use adversarial frames of width 2 pixels: 
in this case we run 100 iterations and 5 restarts of APGD on the margin loss. 
In addition to the models introduced in Sec.~\ref{sec:models}, we train, with 100 epochs of adversarial training wrt $\ell_\infty$ and $\epsilon=4/255$ (details in the appendix), a ConvNeXt-T and a modification of ResNet-50 which includes patchified stem and depth-wise convolutions (see Sec.~\ref{sec:abl_rn50} for details). 
Further, we retrain a ResNet-50 with GELU, as in \citet{bai2021are}, in the same setup as the ConvNeXt-T for further comparison.

\textbf{Results:} 
Table~\ref{tab:rob_linf_models} shows that our trained ConvNeXt has $4.5\%$ higher robust accuracy in $\ell_\infty$ than the XCiT-S (for which significant optimization of hyperparameters has been performed in \citet{debenedetti2022adversarially}) and $9.7\%$ more robust than the ResNet-50. Remarkably, ``ResNet-50 modified'', where patchified stem and depth-wise convolutions have been 
introduced, performs almost as good as the ConvNeXt in $\ell_\infty$. When looking at the generalization to unseen threat models, DeiT-S and XCiT-S attain the best results, especially wrt $\ell_p$-norms. Moreover, ``ResNet-50 modified'' improves most of the time the performance over ResNets and ConvNeXt, but is worse than the transformer-based models. It is an interesting open question if one can combine the advantages of transformers and improved ResNet-architecture further to define an even better architecture for adversarial robustness.

\begin{table*}[h]
    \centering \small
    \tabcolsep=1.4mm
    \caption{Robust accuracy on 1000 points of $\ell_\infty$-robust models to $\ell_p$-bounded 
    and sparse attacks. 
    }
    \label{tab:rob_linf_models}
    \begin{tabular}{C{26mm}| C{25mm} |*{2}{C{9mm}} |*{5}{C{9mm}}}
    \multirow{2}{*}{\textit{model}}&
     \multirow{2}{*}{\textit{reference}}&\multicolumn{2}{c}{\textit{seen}} & \multicolumn{5}{c}{\textit{unseen}} \\
         
         & &\textit{clean}& $\ell_\infty$ 
&$\ell_2$ 
& $\ell_1$ 
& $\ell_0$ & \textit{patches}
& \textit{frames}
\\ \toprule
ResNet-50 & \citet{bai2021are} & 68.2 & 36.7 & 15.6 & 3.1 & 5.7 & 14.5 & 19.7 \\ 
WideResNet-50-2 &\citet{salman2020adversarially}& 69.2 & 38.2 & 20.2 & 4.1 & 5.1 & 16.4 & 19.7 \\ 
DeiT-S & \citet{bai2021are} &66.4 & 35.6 & 40.1 & 21.7 & 24.5 & 21.4 & 25.8 \\ 
XCiT-S & \citet{debenedetti2022adversarially} & 72.8 & 41.7 & 45.3 & 22.2 & 20.8 & 18.7 & 21.6 \\ \midrule
ResNet-50 & retrained & 68.9 & 36.7 & 19.2 & 3.9 & 6.6 & 16.3 & 24.3 \\ 
ResNet-50 modified & retrained & 69.9 & 44.0 & 34.3 & 11.2 & 14.5 & 17.1 & 20.8 \\ 
ConvNeXt-T & retrained & 70.7 & 46.2 & 30.6 & 9.2 & 16.4 & 21.9 & 18.7
\\ \bottomrule
\end{tabular} \end{table*}

\section{From ResNet to ConvNeXt for robustness} \label{sec:abl_rn50}

\begin{table*}[h]
    \centering \small

    \caption{Robust accuracy on 1000 points to $\ell_p$-bounded perturbations 
    of models with different architectures adversarially trained wrt $\ell_\infty$.}
    \label{tab:abl_rn50}

    \begin{tabular}{L{105mm} |*{3}{c} 
    }
         \textit{model} & \textit{clean}&  $\ell_\infty$ 
&$\ell_2$ 
\\ \toprule

\rowcolor{Gray}
ResNet-50 & 60.7 & 28.1 & 16.7 \\ 
\hfill 3:3:9:3 stage ratio & 62.0 & 27.5 & 18.5 \\ 
\hfill ReLU $\rightarrow$ GELU & 61.8 & 29.6 & 13.6 \\ 
\hfill depth-wise conv. with increased width & 63.0 & 28.5 & 19.6 \\ 
\hfill patchify stem & 61.4 & 27.4 & 35.4 \\ 
\hfill patchify stem + depth-wise conv. with increased width & 63.4 & 29.1 & 36.9 \\ 
\hfill patchify stem + GELU & 64.4 & 33.4 & 37.6 \\
\hfill patchify stem + GELU + depth-wise conv. with increased width & 64.6 & 35.0 & 38.2 \\
\midrule
\rowcolor{Gray} ResNet-50 + patchify stem + GELU + depth-wise conv. with increased width & 64.6 & 35.0 & 38.2 \\
\hfill + 3:3:9:3 stage ratio & 66.3 & 35.5 & 39.3 \\ 
\hfill + inverted bottleneck & 66.0 & 33.3 & 28.9 \\ 
\hfill + fewer activations and normalizations & 64.6 & 31.5 & 37.0 \\ 
\hfill + BatchNorm $\rightarrow$ LayerNorm & 62.6 & 34.0 & 39.2 \\ 
\hfill + move downsampling to a separate layer & 64.1 & 34.4 & 37.7 \\ \midrule
ConvNeXt-T without Layer Scale & 65.2 & 36.5 & 36.7 \\ 
\rowcolor{Gray}ConvNeXt-T & 65.2 & 37.9 & 29.5 
\\ \bottomrule
\end{tabular}
\end{table*}

Following the observations of Sec.~\ref{sec:generalization_robustness} we explore which components of the transformers-based architectures might lead i) to better robustness into the threat model used for training ($\ell_\infty$) and ii) better generalization of robustness to unseen threat model, which is relatively weak for ResNets. 
We follow the steps of \citet{liu2022convnet} and start from the basic version of ResNet-50 (as implemented in \torchvision) and progressively update it. For each resulting architecture we train a plain model and use it as initialization for single step adversarial training with $\epsilon=4/255$ in the $\ell_\infty$-threat model (this improves the clean accuracy of robust models with the short training). Given the high computational cost, we use the \ffcv library \citep{leclerc2022ffcv} for preprocessing the dataset, train for 16 epochs with batch size of 2048 (reduced only when necessary to fit a model into GPU memory), cyclic schedule for the learning rate with maximum value of 0.004, AdamW \citep{loshchilov2017decoupled} as optimizer, weight decay of 0.05 (more details in appendix). 
Table~\ref{tab:abl_rn50} tracks the clean accuracy and robust accuracy wrt $\ell_\infty$ and $\ell_2$ (as a proxy for generalization to other threat models), with bounds of 4/255 and 2 respectively, of each model, evaluated with APGD for cross-entropy and DLR loss from AutoAttack.

\textbf{Effect of main architecture components:} We start with testing individually the effect of the main modifications brought by \citet{liu2022convnet}: 1) changing the number of residual blocks in each stage, 2) using GELU as activation function instead of ReLU, 3) using depth-wise convolutions together with increasing of $1.5\times$ the width to roughly preserve the number of parameters, 4) patchified stem. Note that \citet{liu2022convnet} modify the activation function only later on in their road towards ConvNeXt, and it has a limited influence on the clean accuracy. However, \citet{xie2020smooth, bai2021are} have shown that a smooth activation function might have a significant impact on robustness, therefore we include it among the initial main components. Table~\ref{tab:abl_rn50} confirms that using GELU improves $\ell_\infty$-robustness, although it decreases it for $\ell_2$. Conversely, 
the patchified stem notably improves the robustness wrt $\ell_2$ compared to the baseline without modifying that for $\ell_\infty$. Combining the patchified stem and GELU leads to improvements in all statistics, with +5\% of robust accuracy wrt $\ell_\infty$ and almost +20\% in that wrt $\ell_2$. This shows that two simple modifications of the architecture can largely influence the effectiveness of adversarial training. A small improvement in all metrics is achieved further adding the depth-wise convolutions: this yield the ``ResNet-50 modified'' used in Sec.~\ref{sec:generalization_robustness}. 
We conjecture that the patchified stem improves robustness wrt $\ell_2$ because it implicitly performs a sort of dimensionality reduction operating on disjoint subsets of input dimensions. Then optimizing wrt adversarial $\ell_\infty$-perturbations should lead effectively to parameters which work well wrt $\ell_2$ too, since due to the dimensionality reduction the threat models become more comparable (note that it holds $\norm{x}_\infty \leq \norm{x}_2 \leq \sqrt{d}\norm{x}_\infty$ where $d$ is the dimension).

\textbf{From improved ResNet to ConvNeXt:} Starting from this modified version of ResNet-50, we progressively take several further steps to reach the ConvNeXt definition. 
In the second part of Table~\ref{tab:abl_rn50} one observes that the remaining modifications have smaller effect on all statistics, although cumulatively bring additional improvements especially for $\ell_\infty$-robustness. Moreover, we notice that Layer Scale \citep{touvron2021going} has a significant effect on the ConvNeXt, improving $\ell_\infty$-robustness at cost of lower robust accuracy wrt $\ell_2$. 
Overall, this process shows that ConvNeXt is a better suited architecture for adversarial robustness than the original ResNet for $\ell_\infty$-robustness, but might be improved for generalization to unseen threat models.

\section{Conclusion} \label{sec:conclusion}
We have analyzed the effect of several architecture components of modern image classifiers on their robustness against different types of adversarial attacks, and, vice-versa, how robustness via adversarial training modifies the parameters learnt. We observed that some architectures appear better suited for robustness and that small modifications of the design might significantly improve robustness and its generalization to unseen attacks. This opens the possibility of searching for networks which are by construction more robust, which has not been extensively explored so far.

\section*{Acknowledgements}
We acknowledge support from the German Federal Ministry of
Education and Research (BMBF) through the Tübingen AI Center (FKZ: 01IS18039A), the DFG Cluster of Excellence ``Machine Learning – New
Perspectives for Science'', EXC 2064/1, project number 390727645, and by DFG grant 389792660 as part of TRR 248.

\bibliographystyle{icml2022}

\newpage
\appendix
\onecolumn
\section{Inner representations of adversarially trained ViTs}
\setlength{\newl}{.22\columnwidth}
\begin{figure}[h] \centering \small
\tabcolsep=1.2pt
\begin{tabular}{c | c | c}
\textit{orig.} & \textit{plain training} & \textit{adversarial training}\\ \toprule
\includegraphics[height=\newl]{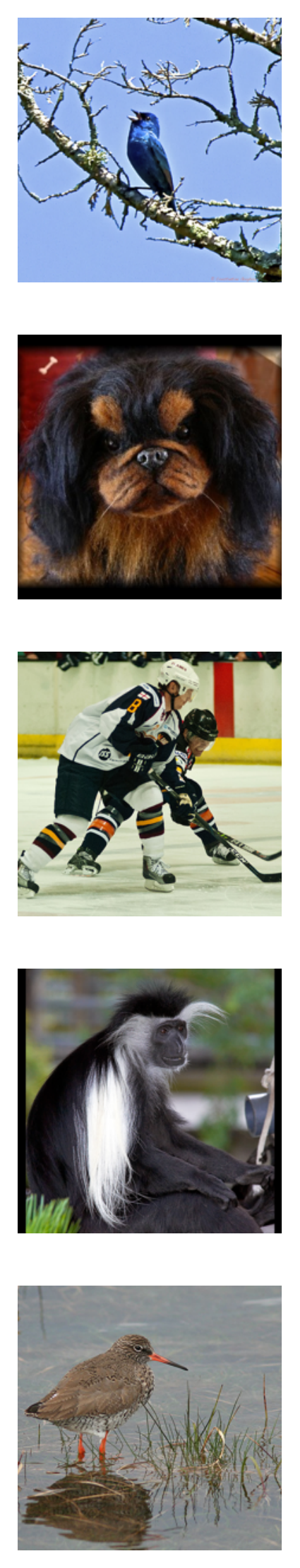}&
\includegraphics[height=\newl]{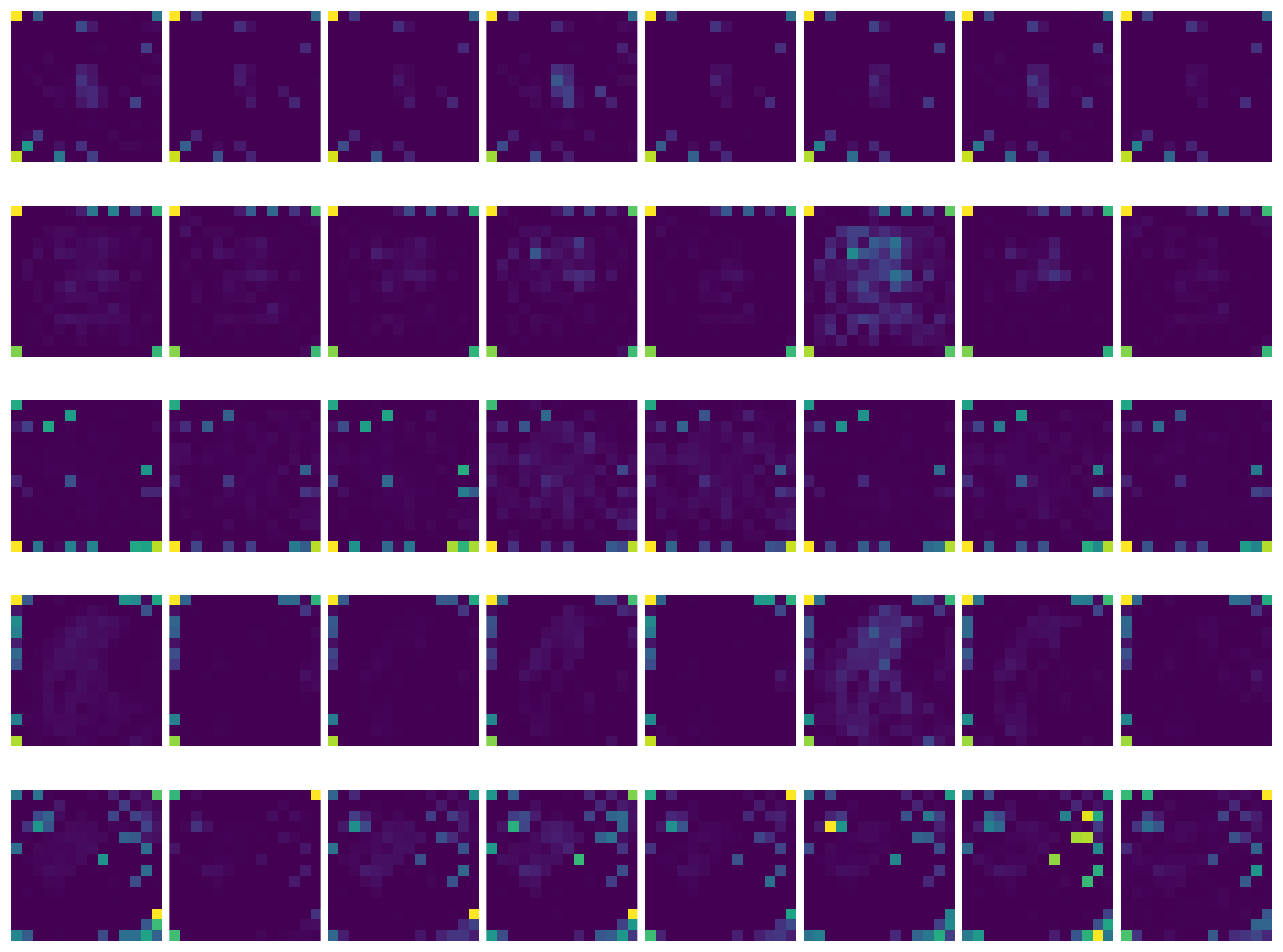}&
\includegraphics[height=\newl]{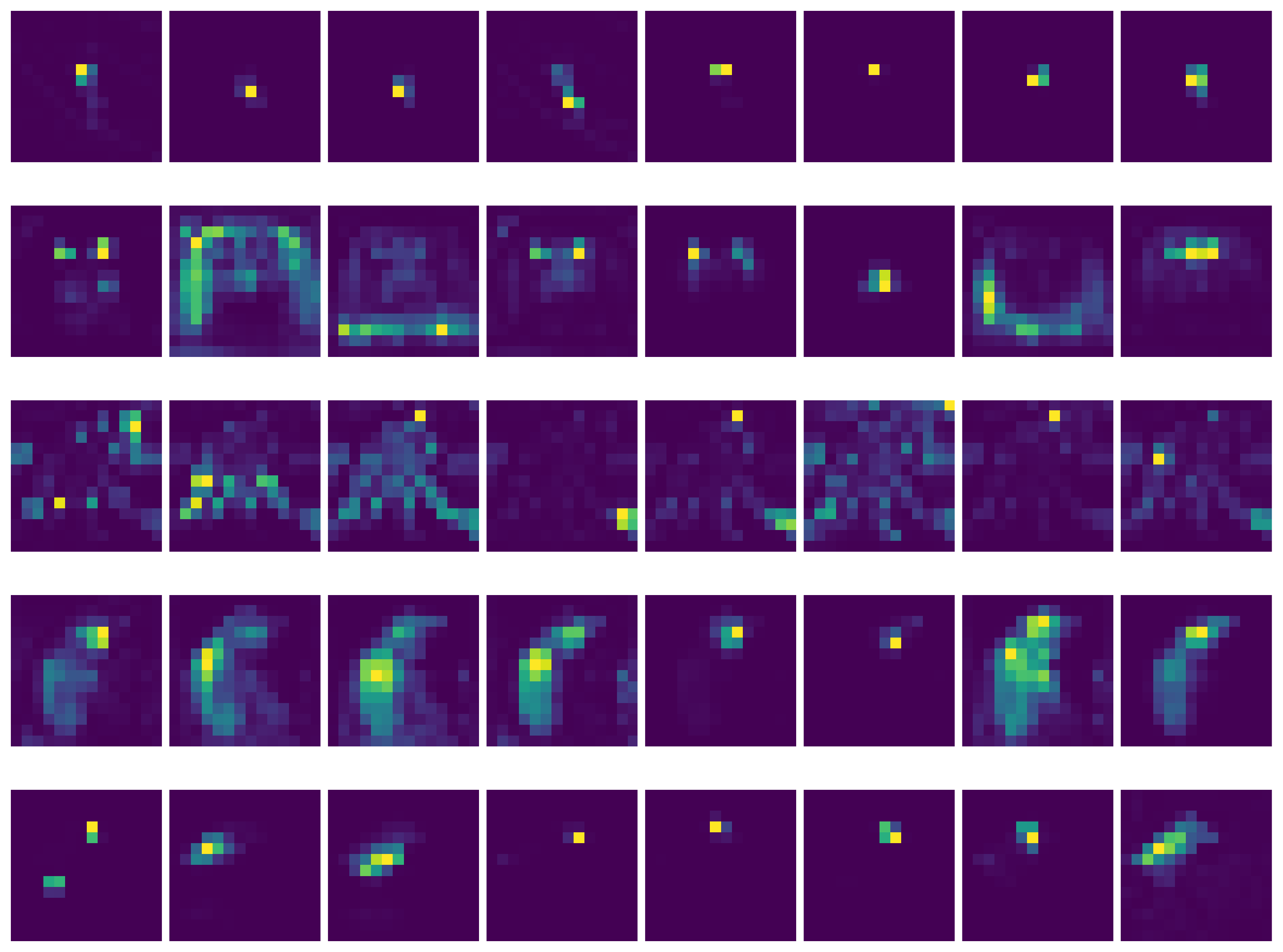}
\end{tabular}
\caption{Attention maps of the \cls token for each head of the last layer of a XCiT-S model: 
in the first column we show the original image, then the 8 maps for the normally trained model, finally the same for the adversarially trained one. Similar to the DeiT-S in Figure \ref{fig:attn_maps}, the attention maps of the robust XCiT-S model are concentrated on the object and highlight different parts of the object.}
\label{fig:attn_maps_xcit}
\end{figure}

\setlength{\newl}{.20\columnwidth}
\begin{figure*}[p] \centering \small
\begin{tabular}{c c}
\includegraphics[height=\newl]{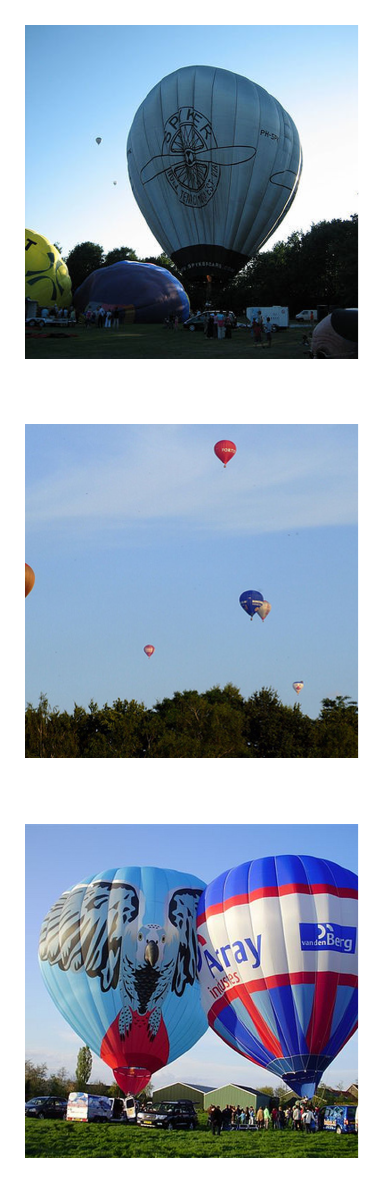} &
\includegraphics[height=\newl]{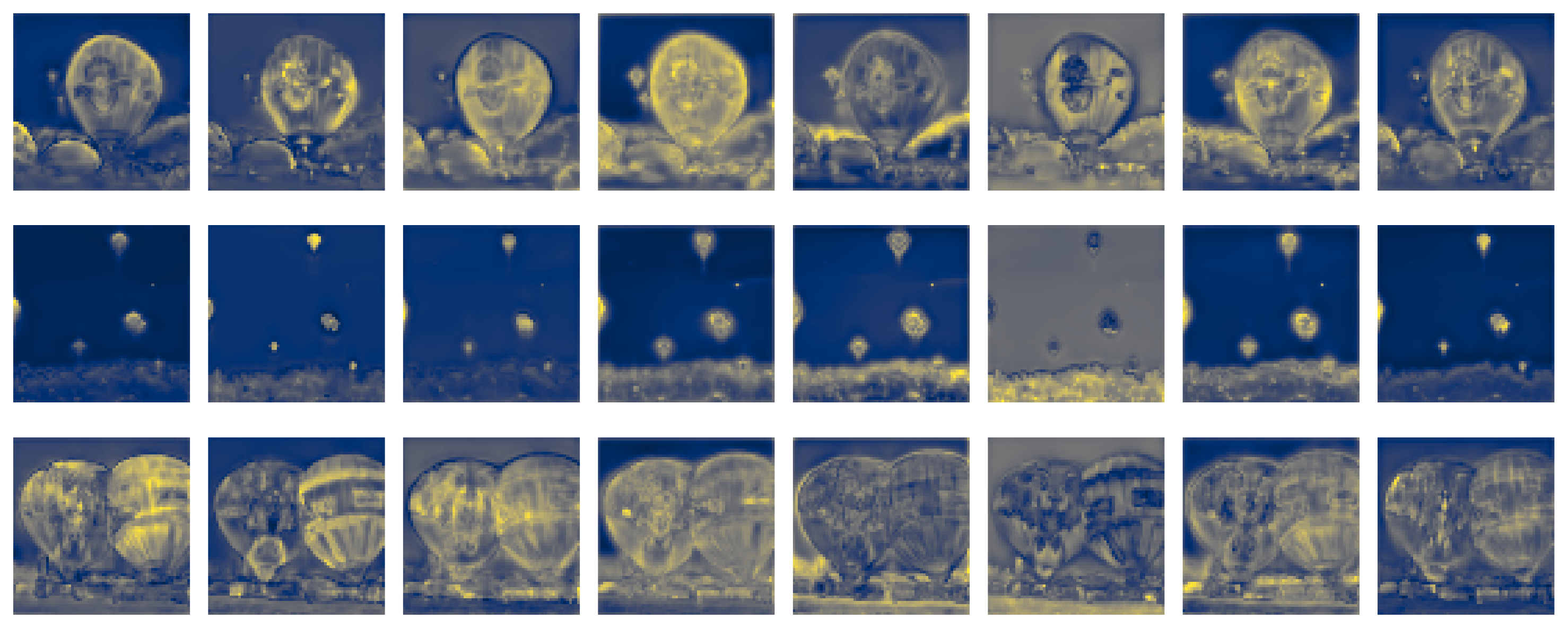} \\[5mm]
\includegraphics[height=\newl]{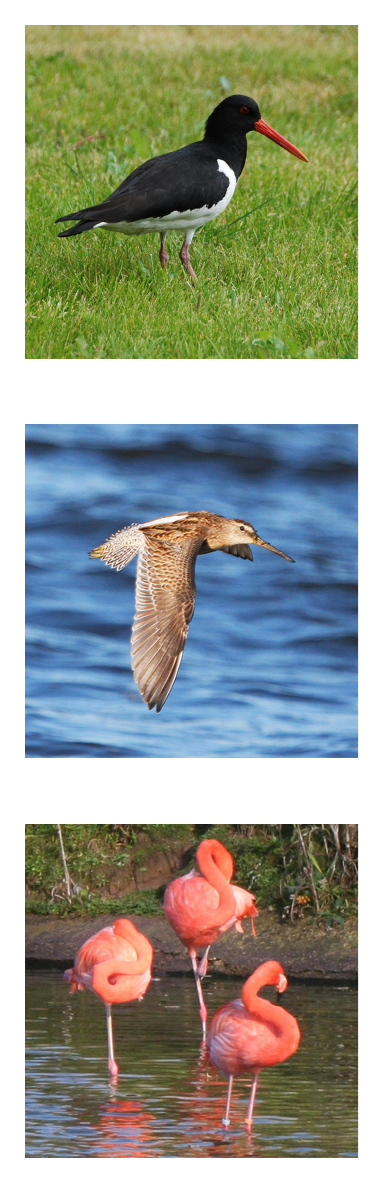} &
\includegraphics[height=\newl]{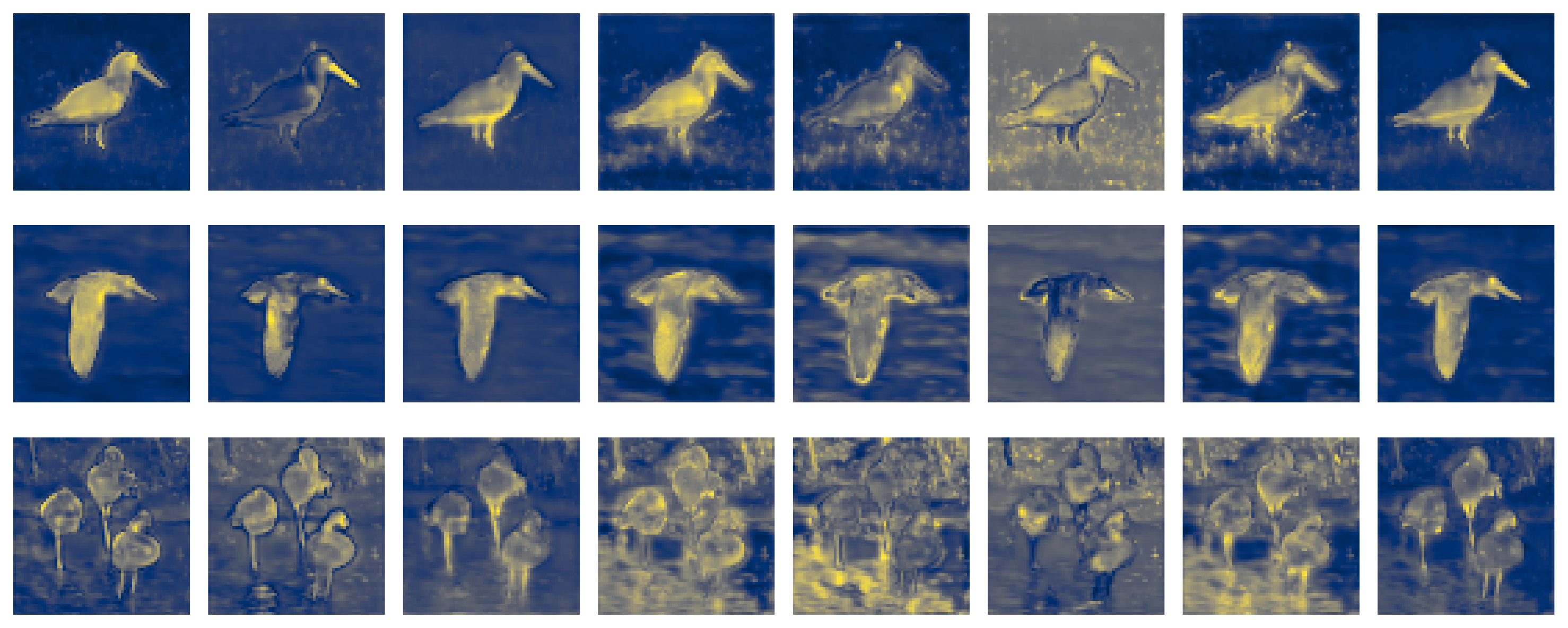}\\[5mm]
\includegraphics[height=\newl]{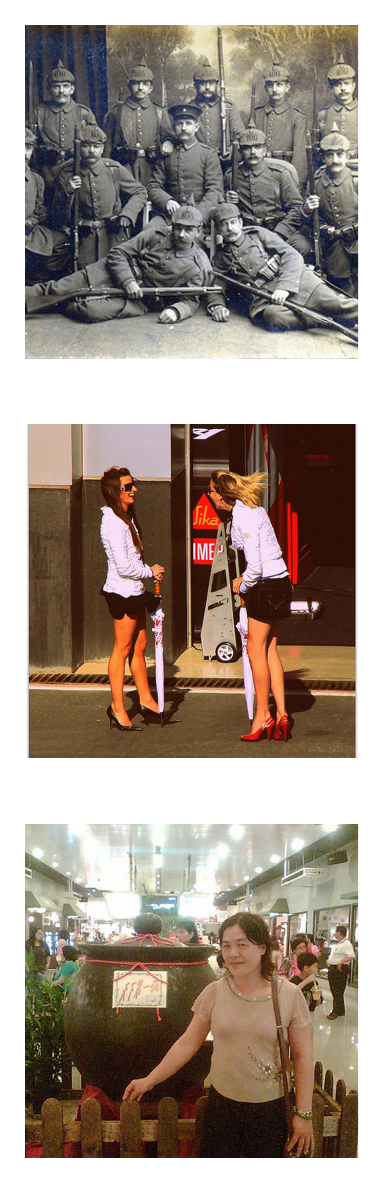} & \includegraphics[height=\newl]{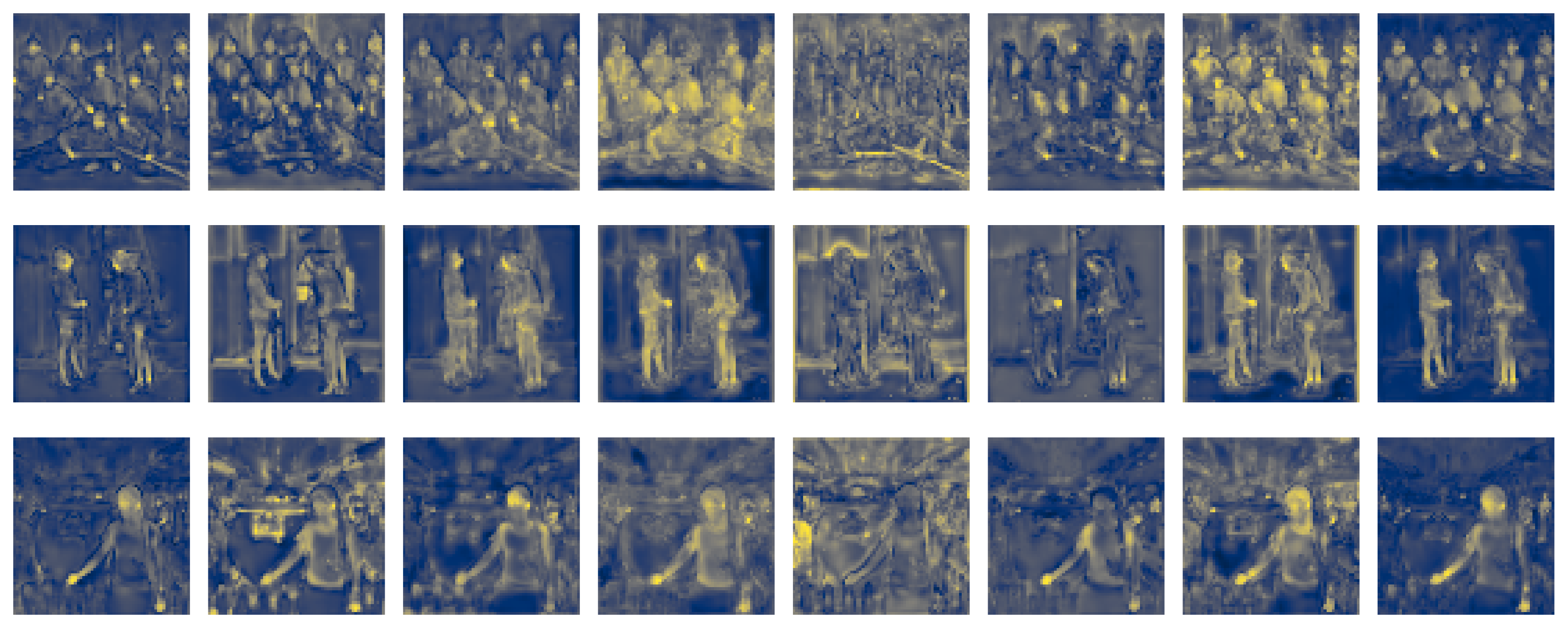}
\end{tabular}
\caption{Visualization of norm across the features dimension of the keys of the last XCA block of the robust XCiT-S for different images of resolution $1024\times 1024$ of similar classes.} \label{fig:app_inner_repr_xcit_classes}
\end{figure*}

\begin{figure*}[p] \centering \small
\begin{tabular}{c c}
\includegraphics[height=\newl]{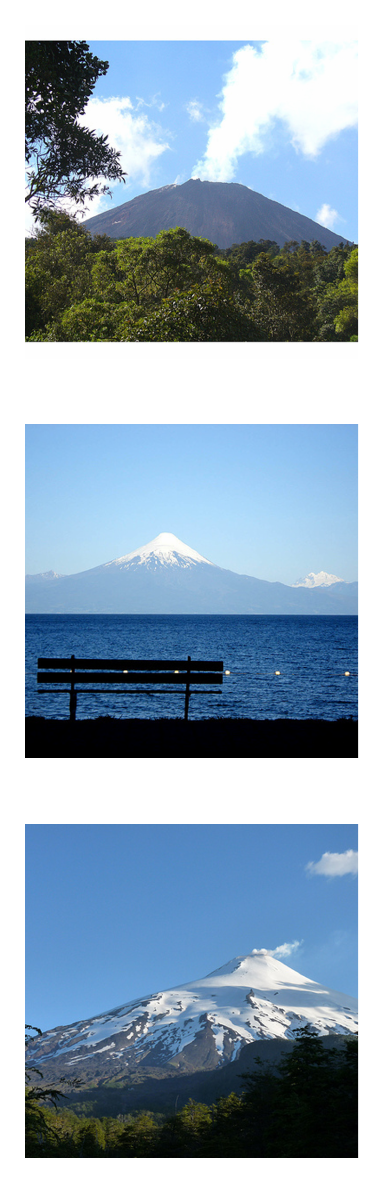} & \includegraphics[height=\newl]{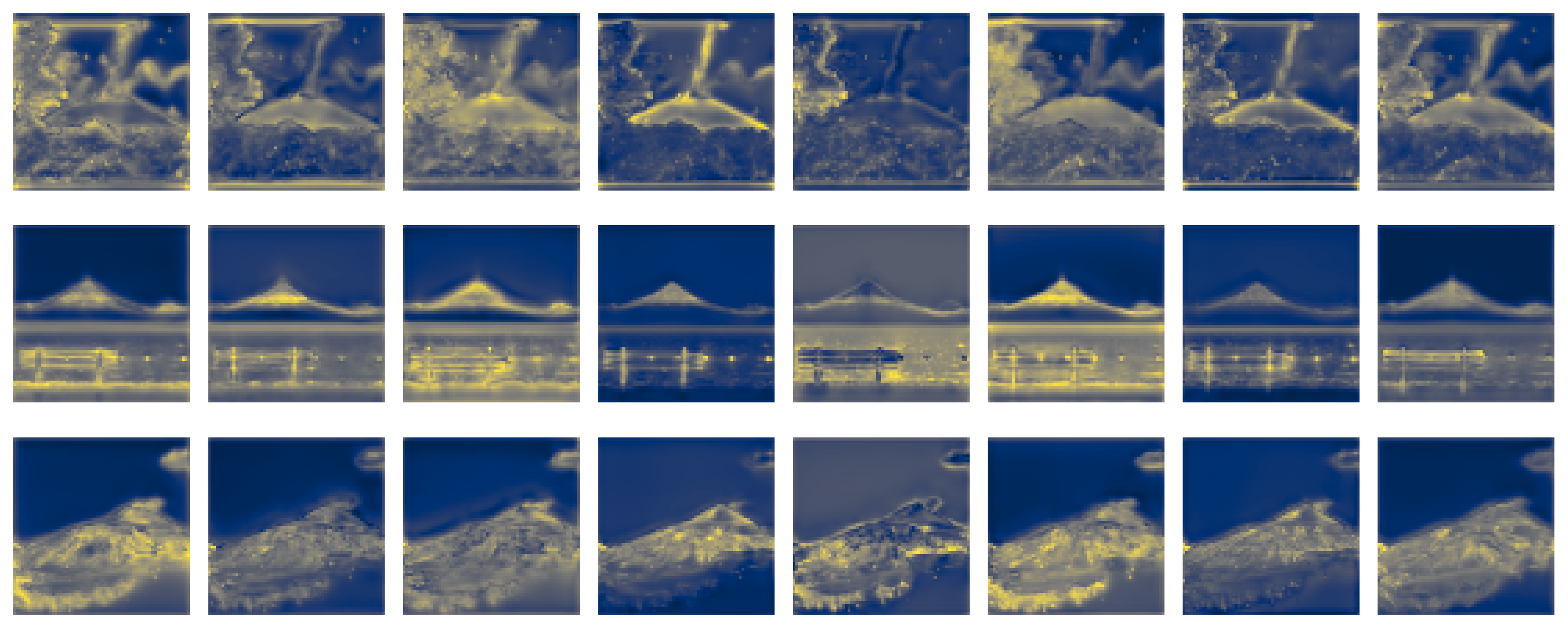}\\[5mm]
\includegraphics[height=\newl]{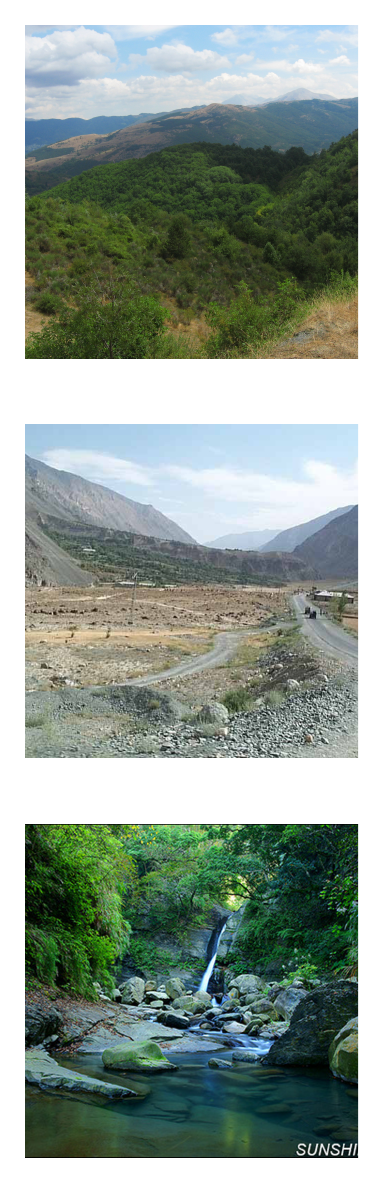}& \includegraphics[height=\newl]{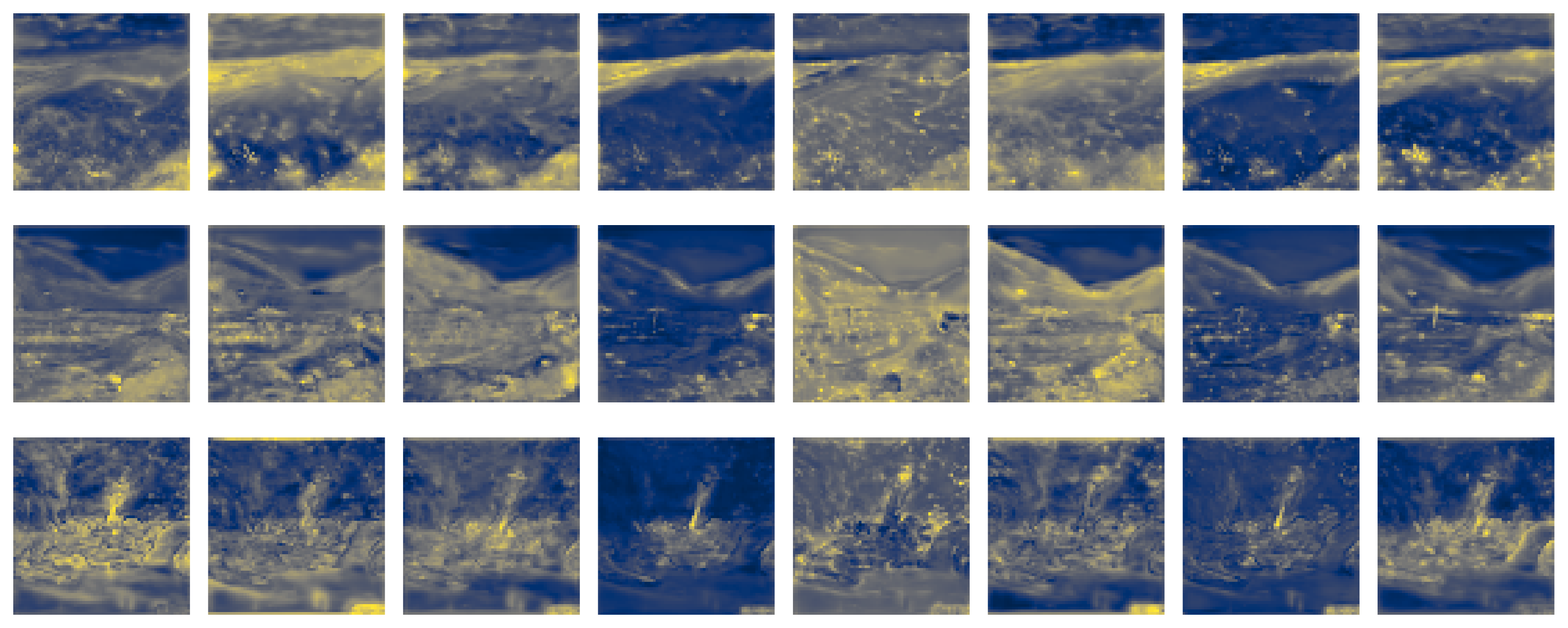}\\[5mm]
\includegraphics[height=\newl]{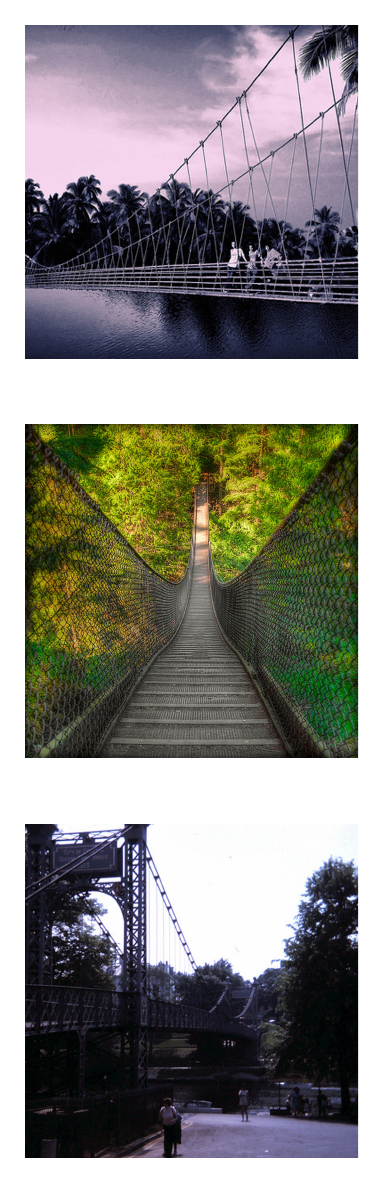} & \includegraphics[height=\newl]{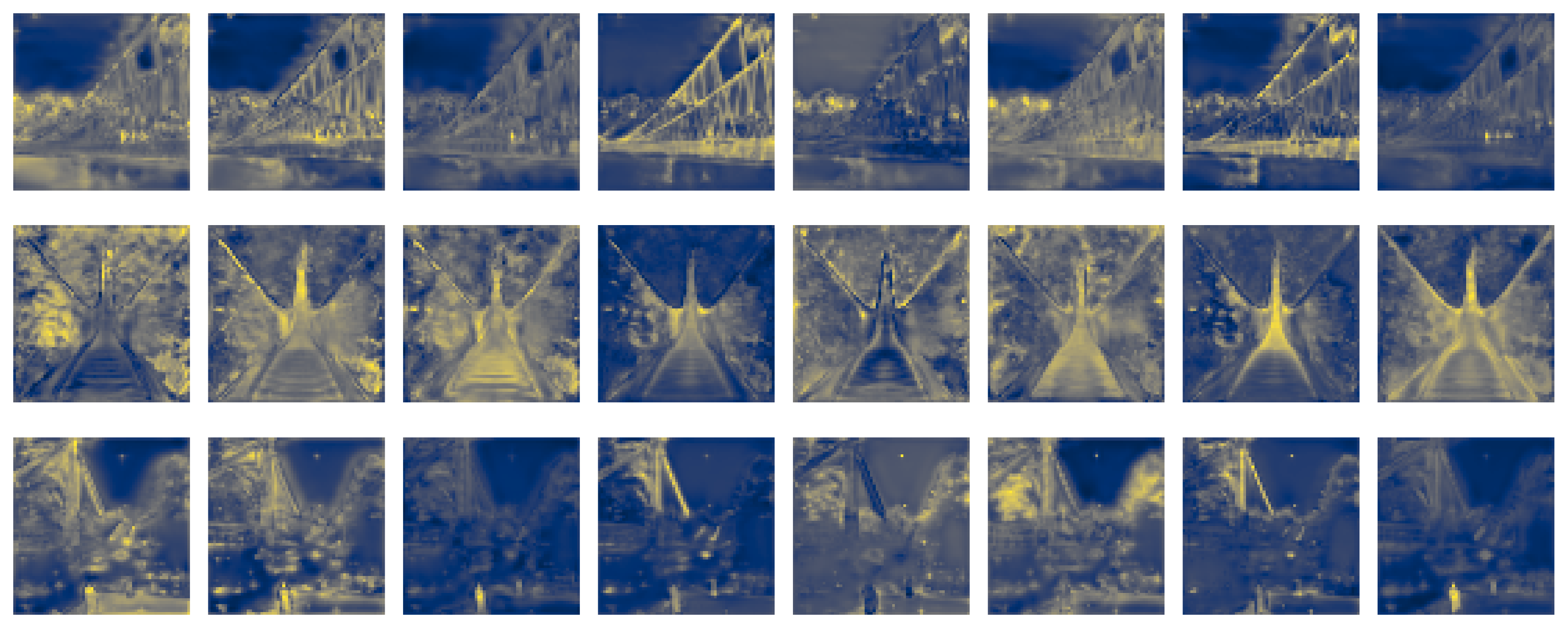}
\end{tabular}
\caption{Visualization of norm across the features dimension of the queries of the last XCA block of the robust XCiT-S for different images of resolution $1024\times 1024$ of similar classes.} \label{fig:app_inner_repr_xcit_classes_queries}
\end{figure*}

As shown in Sec.~\ref{sec:inner_repr_xcit}, it is possible to visualize salient regions of an image with the keys matrices of the XCA blocks of the XCiT models, with adversarially trained models yielding more interpretable maps. We provide further examples of this in Fig.~\ref{fig:app_inner_repr_xcit_classes} with images in resolution $1024\times 1024$ with similar contents. Moreover, we generate similar maps using the queries matrices (instead of keys): Fig.~\ref{fig:app_inner_repr_xcit_classes_queries} shows that even in this case for each head only salient regions are triggered.

\section{Experimental details 
}
In the following we provide additional details about the setup of the various experiments and further results.

\subsection{Generalization of robustness to unseen threat models}
\textbf{Experimental setup:} For the additional classifiers we perform adversarial training wrt $\ell_\infty$ at $\epsilon=4/255$ adapting the pipeline of \url{https://github.com/facebookresearch/deit}. We train for 100 epochs with batch size of 1024, AdamW optimizer, cosine learning rate with maximum value of 0.001 after 10 epochs of warm-up (during which the learning rate is lineraly increased) and 5 of cool-down, weight decay of 0.05. As noted by \citet{debenedetti2022adversarially}, for adversarial training it is not necessary to use heavy augmentation techniques.
Since both the ``ResNet-50 modified'' and ConvNeXt-T suffer from catastrophic overfitting \citep{Wong2020Fast} when using FGSM without random initialization, we prevent it by increasing the number of steps for the inner maximization in adversarial training (up to 3 for former, up to 2 for the latter). Therefore we retrain the standard ResNet-50 with GELU, as in \citet{bai2021are}, with 2 steps 
to match the budget given to ConvNeXt-T. We note that its results are consistent with the ResNet-50 from \citet{bai2021are} trained with single step adversarial training (see Table~\ref{tab:rob_linf_models}). For all architectures we use a plain model, trained for the ablation study of ResNet-50 (see Sec.~\ref{sec:abl_rn50}), as initialization. We select the model among the checkpoints at different epochs as the most robust one to FGSM attack on 5000 images.

\textbf{Additional results:}
We additionally test the robustness of the various models reported in Table~\ref{tab:rob_linf_models} in the $\ell_p$-threat models with the full version of AutoAttack \citep{croce2020reliable} and report the results in Table~\ref{tab:app_rob_linf_models}, where one can see that the robust accuracy decreases at most by 0.3\% by using the missing attacks in AutoAttack, this is FAB- and Square-Attack.

\begin{table}[h]
    \centering \small
    \tabcolsep=1.4mm
    \caption{Robust accuracy on 1000 points of $\ell_\infty$-robust models to $\ell_p$-bounded with AutoAttack.
    }
    \label{tab:app_rob_linf_models}
    \begin{tabular}{C{26mm}| C{25mm} |*{2}{C{9mm}} |*{5}{C{9mm}}}
    \multirow{2}{*}{\textit{model}}&
     \multirow{2}{*}{\textit{reference}}&\multicolumn{2}{c}{\textit{seen}} & \multicolumn{2}{c}{\textit{unseen}} \\
         
         & &\textit{clean}& $\ell_\infty$ 
&$\ell_2$ 
& $\ell_1$ \\ \toprule
ResNet-50 & \citet{bai2021are} & 68.2 & 36.7 & 15.4 & 3.1 \\ 
WideResNet-50-2 & \citet{salman2020adversarially} & 69.2 & 38.2 & 19.9 & 4.1 \\ 
DeiT-S & \citet{bai2021are} & 66.4 & 35.6 & 40.1 & 21.5 \\ 
XCiT-S & \citet{debenedetti2022adversarially} & 72.8 & 41.7 & 45.0 & 22.0 \\ \midrule
ResNet-50 & retrained & 68.9 & 36.7 & 18.9 & 3.9 \\ 
ResNet-50 modified & retrained & 69.9 & 44.0 & 34.2 & 11.2 \\ 
ConvNeXt-T & retrained & 70.7 & 46.2 & 30.6 & 9.2
\\ \bottomrule
\end{tabular} \end{table}

\subsection{From ResNet to ConvNeXt for robustness}
\textbf{Experimental setup:}
For the models reported in Sec.~\ref{sec:abl_rn50} we rely on the \ffcv library \citep{leclerc2022ffcv} since it provides a significant speed-up of training. In particular, we follow the training script, adapted to the adversarial training setup, available at \url{https://github.com/libffcv/ffcv-imagenet}. In particular, we use the standard pre-processing with the exception of the maximum size of the images which we set to 400 pixels instead of 500 (for the short training with only 16 epochs this does not appear to degrade performance). Moreover, we disable BlurPool and test-time augmentation. For the inner maximization process in adversarial training we use FGSM without random initialization.

\textbf{Additional results:} We report in Table~\ref{tab:app_abl_rn50_plain} the clean accuracy (on the 1000 points used for the evaluation of robustness) of the classifiers naturally trained with the various architectures and used as initialization for the robust models shown in Table~\ref{tab:abl_rn50}. We observe that all models have clean accuracy above 70\% and in the same range. Note the short training scheme we employ here is very different from the setup of \citet{liu2022convnet} where ConvNeXt significantly outperforms ResNet.

\begin{table}[h]
    \centering \small
    
    \caption{We report the 
    clean accuracy (on 1000 points) of the models with different architectures 
    obtained with plain training.}
    \label{tab:app_abl_rn50_plain}
\begin{tabular}{L{105mm} | C{12mm}}
\textit{model} & \textit{clean}\\ \toprule
\rowcolor{Gray}
ResNet-50 & 70.9 \\ 
\hfill 3:3:9:3 stage ratio & 71.1 \\ 
\hfill ReLU $\rightarrow$ GELU & 72.6 \\ 
\hfill depth-wise conv. with increased width & 73.2 \\ 
\hfill patchify stem & 72.0 \\ 
\hfill patchify stem + depth-wise conv. with increased width & 73.9 \\ 
\hfill patchify stem + GELU & 72.7 \\ \midrule
\rowcolor{Gray} ResNet-50 + patchify stem + GELU + depth-wise conv. with increased width & 72.5 \\ 
\hfill + 3:3:9:3 stage ratio & 73.9 \\ 
\hfill + inverted bottleneck & 71.0 \\ 
\hfill + fewer activations and normalizations & 71.2 \\ 
\hfill + BatchNorm $\rightarrow$ LayerNorm & 71.8 \\ 
\hfill + move downsampling to separate layer & 72.5 \\ \midrule
ConvNeXt-T without Layer Scale & 72.0 \\ 
\rowcolor{Gray} ConvNeXt-T & 71.8
\\ \bottomrule
\end{tabular}
\end{table}

\section{Visualization of adversarial perturbations}
\citet{bhojanapalli2021understanding} noticed that adversarial perturbations generated for ViTs show a grid structure which resembles that of the tokens. We show in Fig.~\ref{fig:app_diff} the adversarial perturbations (summed over color channels) generated for the $\ell_2$-threat model ($\epsilon=2$ is used)
for each naturally trained classifier. The grid structure appears for the models using input tokenization. 
However, among those, the effect looks 
stronger for DeiT, which does not use any convolutional component, and milder for XCiT, which relies on both convolutional layers and cross-covariance self-attention.

\setlength{\newl}{.55\columnwidth}
\begin{figure}[h] \centering \small
\begin{tabular}{cc} \textit{original} & 
\includegraphics[align=c, width=\newl]{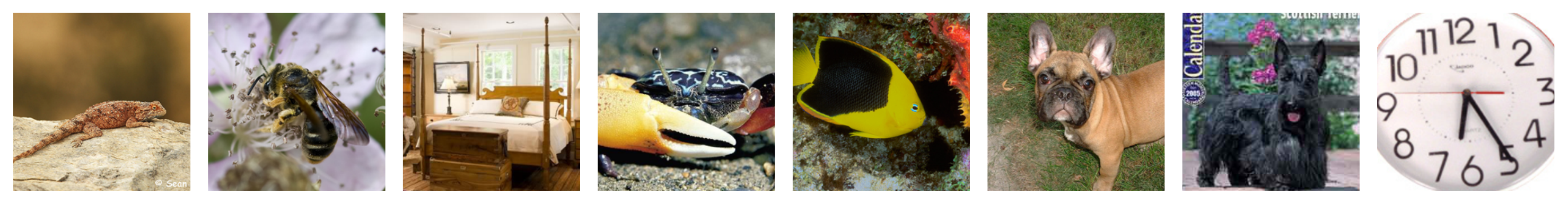}\\
DeiT-S & \includegraphics[align=c, width=\newl]{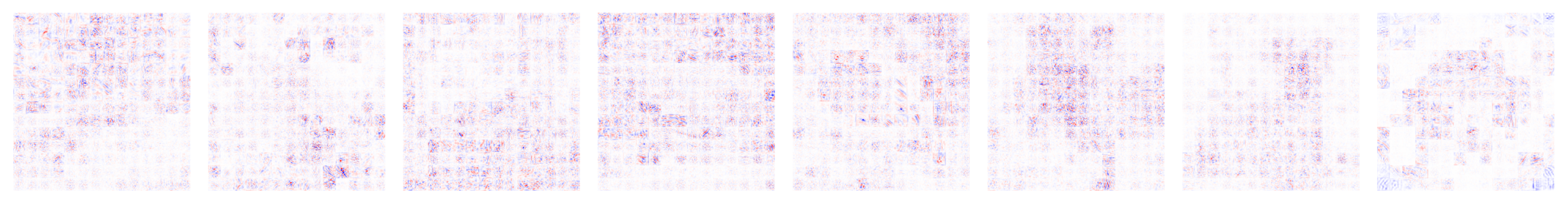}\\
XCiT-S & \includegraphics[align=c, width=\newl]{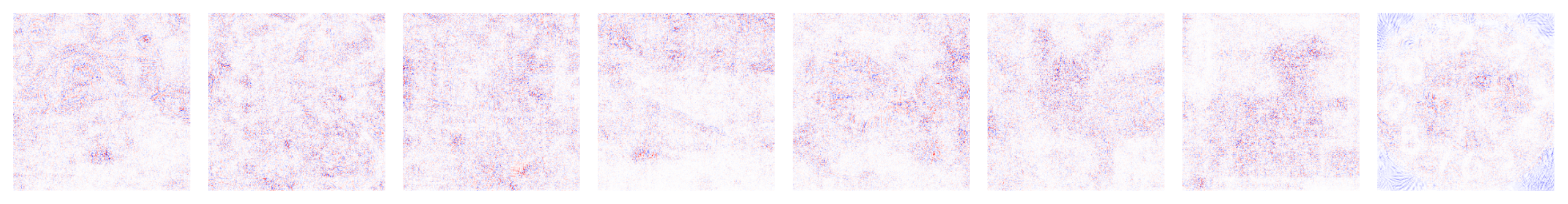}\\
ResNet-50 & \includegraphics[align=c, width=\newl]{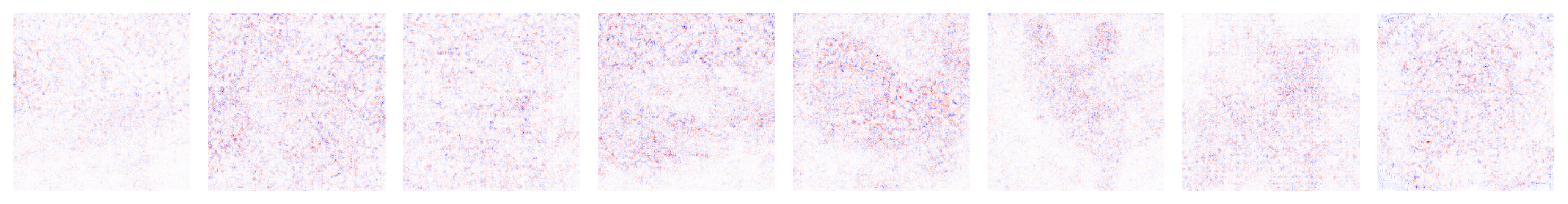}
\end{tabular}
\caption{We plot the adversarial perturbations wrt $\ell_2$ of different natural models: the perturbations are summed over color channels and rescaled so that white areas correspond to zero entries, blue negative, red positive (with intensity indicating their magnitude).} \label{fig:app_diff}
\end{figure}

\end{document}